\documentclass[3p, 11pt]{elsarticle} 
\usepackage[english]{babel}
\usepackage{natbib}
\usepackage[ruled,linesnumbered]{algorithm2e}
\usepackage{amssymb,amsmath,graphicx}
\usepackage{amsthm}
\usepackage{float}
\usepackage{afterpage}
\usepackage{changepage}
\usepackage{dcolumn}
\usepackage{multirow}
\usepackage{color}
\usepackage[pdftex]{pict2e}
\usepackage{enumitem}
\usepackage{longtable}
\usepackage{indentfirst}
\usepackage{mathtools}
\usepackage{subcaption}
\usepackage{tabularx}
\usepackage{changes}
\usepackage{multicol}

\usepackage[figuresright]{rotating}

\usepackage{hyperref}       
\usepackage{url}            
\usepackage{booktabs}       
\usepackage{nicefrac}       
\usepackage{microtype}      
\usepackage{lipsum}
\usepackage{graphicx}       
\usepackage{listings}       
\usepackage{xcolor}         
\usepackage{multirow}       
\usepackage[pdftex]{pict2e}
\usepackage{float}

\SetKwComment{Comment}{\color{green!50!black}// }{}

\SetKwProg{Function}{Function}{}{end}

\allowdisplaybreaks

\definecolor{codegreen}{rgb}{0,0.6,0}
\definecolor{codegray}{rgb}{0.5,0.5,0.5}
\definecolor{codepurple}{rgb}{0.58,0,0.82}
\definecolor{backcolour}{rgb}{0.95,0.95,0.92}

\begin{document}

\begin{frontmatter}




\title{Hybridising Reinforcement Learning and Heuristics for Hierarchical Directed Arc Routing Problems}


\author[1]{Van Quang Nguyen}
\author[2]{Quoc Chuong Nguyen}
\author[3]{Thu Huong Dang}
\author[1]{Truong-Son Hy \texorpdfstring{\corref{cor1}}}

\cortext[cor1]{Correspondence to thy@uab.edu}
\affiliation[1]{organization={University of Alabama at Birmingham},
postcode={AL 35294},
city={Birmingham},
country={The United States}}
\affiliation[2]{organization={State University of New York at Buffalo},
postcode={NY 14260},
city={Buffalo},
country={The United States}}
\affiliation[3]{organization={Lancaster University},
postcode={LA1 4YX},
city={Lancaster},
country={The United Kingdom}}

\address{}

\begin{abstract}
The Hierarchical Directed Capacitated Arc Routing Problem (HDCARP) is an extension of the Capacitated Arc Routing Problem (CARP), where the arcs of a graph are divided into classes based on their priority. The traversal of these classes is determined by either precedence constraints or a hierarchical objective, resulting in two distinct HDCARP variants. To the best of our knowledge, only one matheuristic has been proposed for these variants, but it performs relatively slowly, particularly for large-scale instances (H\`a et al., 2024). In this paper, we propose a fast heuristic to efficiently address the computational challenges of HDCARP. Furthermore, we incorporate Reinforcement Learning (RL) into our heuristic to effectively guide the selection of local search operators, resulting in a hybrid algorithm. We name this hybrid algorithm as the \textbf{H}ybrid \textbf{R}einforcement Learning and Heuristic Algorithm for \textbf{D}irected \textbf{A}rc Routing (HRDA). The hybrid algorithm adapts to changes in the problem dynamically, using real-time feedback to improve routing strategies and solution's quality by integrating heuristic methods. Extensive computational experiments on artificial instances demonstrate that this hybrid approach significantly improves the speed of the heuristic without deteriorating the solution quality. Our source code is publicly available at: \url{https://github.com/HySonLab/ArcRoute}.
\end{abstract}


\end{frontmatter}


\section{Introduction}
\label{sec:intro}
\emph{Vehicle Routing Problems} (VRPs) are an important family of combinatorial optimisation problems, and they have received a great deal of attention from the Operational Research and Optimisation communities \cite{GRW08,TV14}.  \emph{Arc Routing Problems} (ARPs) are a special kind of VRPs, in which the demands are located along the edges or arcs of the network, rather than at the nodes \cite{TD22}. In this paper, we consider the \emph{Hierarchical Directed Capaciated Arc Routing Problem}, or HDCARP for short. In the HDCARP, we are given a strongly connected directed graph \( G = (V, A) \), where $V$ is the vertex set and $A$ is the (directed) arc set. Node \( v_0 \in V \) is called the \emph{depot}. Each arc \( a \in A \) has a positive rational \emph{traversal time} \( d_a \). We are also given a set  \( A_r \subseteq A \) of \emph{required arcs}. Each require arc \( a \in A_r \) has a positive rational \emph{demand} \( q_a \), a positive rational \emph{servicing time} \( s_a \), and a \emph{priority level} \( p_a \). The set \( A_r \) is partitioned into \( p \) pairwise disjoint classes \( A_r^1, A_r^2, \ldots, A_r^p \) such that \( A_r = A_r^1 \cup A_r^2 \cup \ldots \cup A_r^p \) and \( A_r^h \cap A_r^k = \emptyset \) for \( h \neq k \) and \( h, k \in \{1, 2, \ldots, p\} \). Let $P$ denote \( \{1, 2, \ldots, p\} \). There is a fleet of identical vehicles \( M \), each of positive rational \emph{capacity} \( Q \), located at the depot. If a vehicle is used, it must start and end its route at the depot. Every required link must be serviced by exactly one vehicle, and the load of each vehicle must not exceed $Q$ at any time.

The aim is to determine a set of routes for a fleet of vehicles that first minimise the maximum completion time of the first priority class, followed by the second priority class, and so on. For each class, the maximum completion time is the time it takes for all vehicles to service the required arcs in that class. This objective is known as the \textit{hierarchical objective}, as described in \cite{Ca04, PLA08,HDNNL24}. 

In the HDCARP, the order in which classes of arcs are serviced is primarily determined by the objective function, meaning that a solution is considered better if it reduces the time taken to complete tasks in higher-priority classes. This flexibility can improve overall completion time, particularly in cases where a lot of time is spent without servicing (\emph{deadheading}), as arcs from lower-priority classes are allowed to be serviced before those from lower-priority classes. However, in certain applications, a linear precedence constraints may apply, which require that higher-priority classes be completed before lower-priority ones. For example, when dispatching ambulances to respond to calls of varying urgency, life-threatening emergencies must be prioritised and addressed before non-urgent medical transport. For consistency with \cite{HDNNL24}, we refer to the HDCARP variant with a flexible servicing order as HDCARP-U, and the one with a strict order imposed by precedence constraints as HDCARP-P.

To the best of our knowledge, only one matheuristic has been proposed for both HDCARP variants in \cite{HDNNL24}, but it is rather slow. This motivates us to develop a fast heuristic for the HDCARP. Constructive heuristics are efficient at quickly generating solutions, but the solution quality is often poor. To address this, local search techniques are commonly applied to improve these initial solutions (e.g., \cite{Pr15, Co21, MP17}). However, since each neighbourhood is of exponential size, we propose integrating \emph{Reinforcement Learning} (RL) into our heuristic to guide the selection of promising neighbourhoods, thereby helping to reduce the computational time of the local search procedure. We will call the  resulting hybrid algorithm for the HDCARP as \textbf{H}ybrid \textbf{R}einforcement Learning and Heuristic Algorithm for \textbf{D}irected \textbf{A}rc Routing (HRDA).


Reinforcement Learning (RL) excels at learning dynamic decision-making strategies by directly interacting with the environment, enabling it to adapt to complex and evolving problem structures. Its ability to balance exploration and exploitation makes it particularly effective for optimizing solutions in combinatorial optimization tasks like routing problems. Standalone RL techniques frequently encounter challenges related to stability and convergence. Our proposed approach harnesses the adaptive learning capabilities of RL, allowing for continuous refinement of routing strategies based on real-time interactions with the environment. This is further enhanced by integrating heuristic optimizations, which improve solution quality and bolster stability. By effectively balancing exploration—where algorithms like Proximal Policy Optimization (PPO) \cite{schulman2017proximalpolicyoptimizationalgorithms} demonstrate superiority—and exploitation, which directs the RL agent toward promising solutions, our framework accelerates convergence and optimizes routing efficiency. Additionally, by strategically reducing the search space and implementing targeted local adjustments, we enhance both the robustness and reliability of the routing solutions.

Our contributions can be summarised as follows:
\begin{itemize}
    \item Developed a hybrid algorithm combining RL with a heuristic approach to address the HDCARP, leveraging RL for adaptive learning and a heuristic approach for efficiently finding high-quality solutions within a short running time.
    \item Designed the framework to significantly accelerate convergence and reduce computational time by dynamically guiding the selection of local search operators to reduce the search space.
    \item Developed metaheuristics for both HDCARP variants.
    \item Conducted extensive computational experiments on artificial HDCARP instances to evaluate the performance of our hybrid algorithm. The results demonstrate that the hybrid algorithm outperforms independent RL and heuristic, as well as proposed metaheuristics, in terms of both solution quality and running time.
\end{itemize}

Our paper is structured as follows. Section \ref{sec:review} gives a brief overview of the relevant literature. Sections \ref{sec:Heuristic} and \ref{sec:methodology} describe our heuristic and hybrid algorithm. Section \ref{sec:meta} presents our metaheuristics. Section \ref{sec:Experimental} presents our computational results, and some concluding remarks are made in Section \ref{sec:conclusion}.

\section{Literature Review}
\label{sec:review}
Since the literature on ARPs is vast, we focus here only on works of direct relevance. Subsection \ref{subse:carp} and \ref{subse:arpsHierarchy} cover CARP and relevant ARPs with precedence relations, while Subsection \ref{subse:LC} and \ref{subse:ML} discuss local search and hybrid algorithms that integrate \emph{Machine Learning} (ML). For more on ARPs in
general, see the books \cite{CL15a,Dr00} and the surveys \cite{Co21,MP17}. The integration of Machine Learning (ML), particularly reinforcement learning and graph neural networks, has seen significant advances in addressing combinatorial optimization problems (COP), including routing. Recent surveys highlight the developments and applications of ML techniques in this domain, see \cite{darvariu2024graphreinforcementlearningcombinatorial, 10263956, 9125934}.

\subsection{Capacitated arc routing problems} \label{subse:carp}

\citet{GW81} introduced the \emph{Capacitated ARP} (CARP) and demonstrated that it is NP-hard in the strong sense. Its directed variant, the DCARP, is also NP-hard, as each edge in the undirected version can be replaced by a pair of anti-parallel arcs. Since the DCARP is a special case of the HDCARP, where $p=1$ and $s_a = t_a$ for all $a \in A_r$, it means that the HDCARP is also NP-hard. 

Exact methods for CARPs are limited to instances with around 180 required edges, as detailed in \cite{BBI15}. For larger instances, many heuristics have been proposed; see \cite{CHG16,Pr15,UMM13,WL18}. Among the many heuristics, we mention greedy constructive heuristics, which remain a crucial area of research for solving the CARPs. It starts with an empty or partial solution and gradually builds a complete solution. At each step, it adds an additional required arc based on factors like distance, cost, time, or even randomly, while ensuring that the vehicle's capacity constraints are satisfied.

In recent years, many approaches for solving CARPs have been based on metaheuristics. For example, Hertz et al. \cite{HLM00}, Brand\~ao and Eglese \cite{BRANDAO20081112}, and Greistorfer \cite{Gr03} introduced different Tabu Search algorithms to solve CARP. Li \cite{Li92}, Eglese \cite{Eg94}, and W{\o}hlk \cite{Wo05} proposed Simulated Annealing algorithms, while Lacomme et al. \cite{LPR01, LPR04} presented memetic algorithms. Santos et al. \cite{SCC10} developed the Ant Colony Optimization. Usberti et al. (2013) \cite{UFF13} proposed a Greedy Randomized Adaptive Search Procedure with Path-Relinking. Additionally, Liu et al. \cite{LJG13} introduced a hybrid algorithm combining a Genetic Algorithm with an Iterated Local Search framework.


\subsection{ARPs with precedence relations} \label{subse:arpsHierarchy}

\citet{Dror1987} introduced the \emph{Hierarchical Chinese Postman Problems} (HCPP). The \textit{Chinese Postman Problems} (CPPs) are a special kind of ARPs, in which all edges should be traversed at least once by a shortest closed walk \cite{Ed65, EGL95a, Gu62}. The HCPP is a variant of CPP, where  edges are grouped into priority classes, and a linear precedence constraint is imposed. While the CPPs can be solved in polynomial time \cite{Ch73, EJ73}, the HCPP is generally NP-hard \cite{Dror1987,ABT21}. Exact methods for the HCPP and its special cases are presented in \cite{Dror1987, GI20, Ca04, KV06, SD15, ABT21}, while heuristics for larger-scale instances can be found in \cite{LG84,CY18}.

\citet{AL88} introduced the directed version of HCPPs. The authors introduced general precedence relations, where all arcs in a high-priority class must be serviced before those in a low-priority class, while arcs in a medium-priority class may be serviced either before or after certain high- or low-priority arcs. They proposed a three-phase constructive heuristic, with each phase focusing on connecting classes, balancing non-symmetric nodes, and identifying feasible routes, respectively. \citet{PLA08} introduced the mixed and multi-vehicle version of HCPPs, where both edges and arcs can be present, and multiple vehicles with unlimited capacity are assumed. The authors introduced two constructive heuristics that decompose the original problem into sub-problems. For more on HCPPs, see the survey\cite{PLC07a, PLC07b}.

\citet{QTL15} introduced the HRPP with multiple vehicles, an extension of the HCPP with multiple vehicles, where only a subset of edges is required to be serviced. The authors considered general precedence relations and proposed an adaptive large neighborhood search metaheuristic for large-scale instances. The mixed version of HRPPs are also studied in \cite{PALC06, CCMPS17a, CCMPS17b, QLLPT17}.

\subsection{Local search procedures} \label{subse:LC}
As one might expect, local search has been widely applied to the CARP. Two kinds of neighbourhoods have proven to be particularly effective:
\begin{itemize}
\item ``swap'' or ``interchange'', which involves exchanging the positions of two arcs within one route or between two routes (e.g., \cite{BEULLENS2003629, 10.1007/978-3-540-28646-2_48});
\item ``shift'' or ``insertion'', which selects an arc and changes its position within the same route or another route (e.g., \cite{BEULLENS2003629, 10.1007/978-3-540-28646-2_48}).
\end{itemize}

One can see that the ``shift'' neighbourhood is a subset of the ``swap'' neighbourhood where one of the arcs involved in the swap is allowed to be empty. To determine whether the ``swap'' or ``shift'' neighbourhood is preferred, one could evaluate their performance in terms of solution quality, computational efficiency, or adaptability to different CARP variants, considering factors such as problem size and complexity.

Some more complicated neighbourhoods include ``general shift'', which relocates an entire sequence of consecutive arcs to a different position within the same route or to a specific position on another route, and ``general swap'' operator, which involves removing two specific sequences of consecutive arcs, either from the same route or from different routes, and then swapping their positions. For brevity, we omit further details. (e.g., \cite{BEULLENS2003629, 10.1007/978-3-540-28646-2_48}).

\subsection{Machine Learning for ARPs/VRP} \label{subse:ML}

Besides the classic techniques used to solve optimization problems, The integration of more advanced techniques, Machine Learning (ML) in this case, into metaheuristics has seen growing interest \cite{BENGIO2021405, Song2019, Talbi2016}, particularly in enhancing the efficiency and efficacy of heuristic approaches for combinatorial optimization problems. Studies like those by \citet{KARIMIMAMAGHAN2022393} and \citet{10.1145/3459664} have reviewed the landscape of ML-augmented metaheuristics, illustrating how supervised and unsupervised learning can be employed to fine-tune heuristic components, adapt parameters dynamically, and predict promising regions of the solution space, thereby significantly improving performance. The fusion of Machine Learning with metaheuristics has also led to the development of novel hybrid methods that leverage the strengths of both domains. For instance, \citet{SONG2024101517} demonstrated how reinforcement learning could be integrated into genetic algorithms to guide the search process more effectively. These hybrid approaches have been shown to outperform traditional metaheuristics in various complex optimization scenarios, underlining the potential of ML techniques to advance the field further.

Specifically, in the context of CARP, several studies have explored the use of Machine Learning to enhance solution methods \cite{10.5555/3295222.3295382, Peng2021} for such optimizations on graphs. For example, \citet{10.5555/3295222.3295382} proposed a reinforcement learning-assisted metaheuristic that optimizes the selection of candidate edges to be traversed, significantly reducing computational time and improving solution quality by combining the Graph Neural Network (GNN) \cite{10.5555/3305381.3305512} model and the Q-Learning algorithm \cite{10.5555/3312046} to generate the approximated solution. This approach has opened new avenues for solving CARP \cite{8790295, RAMAMOORTHY2024200300, guo2024efficientlearningbasedsolvercomparable}, demonstrating that ML can be a powerful tool in addressing the inherent challenges of such complex optimization problems.

\section{Heuristic algorithm}
\label{sec:Heuristic}
In this section, we present a fast heuristic for both HDCARP variants. The heuristic has two ``phases'', which are described in the following subsections.

\subsection{Construction phase}
To compute an initial solution, we developed a greedy constructive heuristic. The heuristic inserts each priority class sequentially. For each priority class, arcs are evaluated one by one. For each arc, the best feasible insertion place across all routes is determined, and routes with feasible insertions are added to the set $M_{\text{feasible}}$. The set $M_{\text{feasible}}$ is then evaluated using a softmax probability distribution, where the scores represent the increase in the completion time of the priority class after inserting the arc into routes and a route in $M_{\text{feasible}}$ is selected randomly based on this distribution. We remark that both variants, HDCARP-U and HDCARP-P, use the same algorithm for the construction phase. However, in HDCARP-U, arcs can be serviced before higher priority classes, which increases the number of feasible insertion places considered for each arc, compared to HDCARP-P.

\begin{algorithm}
    \caption{Greedy Constructive Heuristic}
    \label{algo:NIH}
    \SetKwInOut{Input}{Input}
    \SetKwInOut{Output}{Output}

    \Input{
        $M$: set of vehicles; $A_r$: set of required arcs; \\
        $A_r^p$: required arcs in class $p$; $\mathcal{D}$: adjacency matrix.
    }
    \Output{Set of routes $\mathcal{R}$}

    \SetKwFunction{ConstructiveHeuristic}{ConstructiveHeuristic}
    \SetKwFunction{GetArcsWithPriority}{GetArcsWithPriority}
    \SetKwFunction{RandomChoiceWithProbability}{RandomChoiceWithProbability}
    \SetKwFunction{Softmax}{Softmax}
    \SetKwFunction{CalcDemand}{CalcDemand}
    \SetKwFunction{CalcObjFunc}{CalcObjFunc}
    \BlankLine

    \Function{\ConstructiveHeuristic{}}{
        \tcp{Initialize routes for each vehicle}
        \ForEach{$m \in M$}{
            $\mathcal{R}[m] \leftarrow [ \ ]$
        }
        \BlankLine
        \tcp{Construct initial solution}
        \ForEach{priority class $p \in \mathcal{P}$}{
            $A_p \leftarrow$ \GetArcsWithPriority{$p$}\;
            \ForEach{arc $a \in A_p$}{
                $M_{\text{feasible}} \leftarrow \emptyset$\;
                $\text{costs} \leftarrow \emptyset$\;
                \ForEach{vehicle $m \in \mathcal{M}$}{
                    $r \leftarrow \mathcal{R}[m] \oplus [a]$\;
                    \If{\CalcDemand{$r,\ \mathcal{A}$} $\leq$ capacity}{
                        $M_{\text{feasible}} \leftarrow M_{\text{feasible}} \cup \{ m \}$\;
                        $\text{costs}[m] \leftarrow$ \CalcObjFunc{$r,\ \mathcal{A},\ \mathcal{D}$}\;
                    }
                }
                \If{$M_{\text{feasible}} \neq \emptyset$}{
                    $m_{\text{choice}} \leftarrow$ \RandomChoiceWithProbability{$M_{\text{feasible}},\ \text{costs}$}\;
                    $\mathcal{R}[m_{\text{choice}}] \leftarrow \mathcal{R}[m_{\text{choice}}] \oplus [a]$\;
                }
            }
        }
        \Return{$\mathcal{R}$}\;
    }

    \Function{\RandomChoiceWithProbability{$M_{\text{feasible}},\ \text{costs}$}}{
        \tcp{Convert costs to probabilities using softmax function}
        $p \leftarrow$ \Softmax{$\text{-costs}$}\;
        
        $m_{\text{choice}} \leftarrow$ \text{Randomly select from $M_{\text{feasible}}$ with probabilities $p$}\;
        
        \Return{$m_{\text{choice}}$}\;
    }
\end{algorithm}

\subsection{Local search phase}
\label{algo:localsearch}
The initial solution is improved using local search procedures in Algorithms \ref{algo:intraP}-\ref{algo:interU}. These procedures apply "swap" neighbourhoods within a route (\texttt{BestSwapIntra}) and between routes (\texttt{BestSwapInter}). For the HDCARP-P variant, arcs are swapped only within the same class (\texttt{GetSubTourP}), while the HDCARP-U variant allows arcs from different classes (\texttt{GetSubTourU}), ensuring its solution is at least as good as HDCARP-P.

Checking the swapping of arcs involves checking vehicle constraints and evaluating the objective function value.
Since the checks can be performed independently for different pairs, we use parallel computing to execute these checks simultaneously, thereby reducing the overall computation time.



\begin{multicols}{2}
\begin{algorithm}[H]
    \caption{Intraroute Swap Operator (HDCARP-P)}
    \label{algo:intraP}
    \SetKwInOut{Input}{Input}
    \SetKwInOut{Output}{Output}

    \Input{
        $r$: a route in solution $\mathcal{R}$; \\
        $A_r^p$: required arcs in class $p$;\\
        $\mathcal{D}$: adjacency matrix.
    }
    \Output{Route $r$}

    \SetKwFunction{Swap}{Swap}
    \SetKwFunction{BestSwapIntra}{BestSwapIntra}
    \SetKwFunction{IntraRouteOptP}{IntraRouteOptP}
    \SetKwFunction{CalcObjFunc}{CalcObjFunc}
    \SetKwFunction{GetSubTourP}{GetSubTourP}

    \Function{\IntraRouteOptP{$r, p$}}{
        $r \leftarrow$ \GetSubTourP{$r, p$}\; 
        $\texttt{Improvement} \leftarrow \texttt{true}$\;
        \While{\texttt{Improvement} = \texttt{true}}{
            $\Delta_{best}, (a_1, a_2) \leftarrow$ \BestSwapIntra{$r$}\;
            
            \If{$\Delta_{best} \ge 0$}{
                 $\texttt{Improvement} \leftarrow \texttt{false}$\;
            }
            \uElse{
            $r \leftarrow$ \Swap{$r, a_1, a_2$}\;
            }
        }
        \Return{Route $r$}\;
    }

\end{algorithm}
\begin{algorithm}[H]
    \caption{Intraroute Operator (HDCARP-U)}
    \label{algo:intraU}
    \SetKwInOut{Input}{Input}
    \SetKwInOut{Output}{Output}
    \Input{
        $r$: a route in solution $\mathcal{R}$; \\
        $A_r^p$: required arcs in class $p$;\\
        $\mathcal{D}$: adjacency matrix.
    }
    \Output{Optimized route $r$}

    \SetKwFunction{Swap}{Swap}
    \SetKwFunction{BestSwapIntra}{BestSwapIntra}
    \SetKwFunction{IntraRouteOptU}{IntraRouteOptU}
    \SetKwFunction{GetSubTourU}{GetSubTourU}

    \Function{\IntraRouteOptU{$r, p$}}{
        $r \leftarrow$ \GetSubTourU{$r, p$}\; 
         $\texttt{Improvement} \leftarrow \texttt{true}$\;
        \While{\texttt{Improvement} = \texttt{true}}{
            $\Delta_{best}, (a_1, a_2) \leftarrow$ \BestSwapIntra{$r$}\;
            
            \If{$\Delta_{best} \geq 0$}{
                $\texttt{Improvement} \leftarrow \texttt{false}$\;
            }\uElse{
            $r \leftarrow$ \Swap{$r, a_1, a_2$}\;
            }
        }
        \Return{Optimized route $r$}\;
    }
\end{algorithm}
\columnbreak
\begin{algorithm}[H]
    \caption{Interroute Swap Operator (HDCARP-P)}
    \label{algo:interP}
    \SetKwInOut{Input}{Input}
    \SetKwInOut{Output}{Output}

    \Input{
        $r_1, r_2$: routes in solution $\mathcal{R}$; \\
        $A_r$: set of required arcs;\\
        $\mathcal{D}$: adjacency matrix.
    }
    \Output{Optimized routes $r_1, r_2$}

    \SetKwFunction{Swap}{Swap}
    \SetKwFunction{BestSwapInter}{BestSwapInter}
    \SetKwFunction{InterRouteOptP}{InterRouteOptP}
    \SetKwFunction{GetSubTourP}{GetSubTourP}

    \Function{\InterRouteOptP{$r_1, r_2, p$}}{
        $r_1 \leftarrow$ \GetSubTourP{$r_1, p$}\;
        $r_2 \leftarrow$ \GetSubTourP{$r_2, p$}\;
         $\texttt{Improvement} \leftarrow \texttt{true}$\;
        \While{\texttt{Improvement} = \texttt{true}}{
            $\Delta_{best}, (a_1, a_2) \leftarrow$ \BestSwapInter{$r_1, r_2$}\;
            
            \If{$\Delta_{best} \geq 0$}{
                $\texttt{Improvement} \leftarrow \texttt{false}$\;
            }\uElse{
            \Swap{$r_1, r_2, a_1, a_2$}\;
            }
        }
        \Return{Optimized routes $r_1, r_2$}\;
    }
\end{algorithm}
\begin{algorithm}[H]
    \caption{Interroute Swap Operator - (HDCARP-U)}
    \label{algo:interU}
    \SetKwInOut{Input}{Input}
    \SetKwInOut{Output}{Output}

    \Input{
        $r_1, r_2$: routes in solution $\mathcal{R}$; \\
        $A_r^p$: set of required arcs;\\
        $\mathcal{D}$: adjacency matrix.
    }
    \Output{Optimized routes $r_1, r_2$}

    \SetKwFunction{Swap}{Swap}
    \SetKwFunction{BestSwapInter}{BestSwapInter}
    \SetKwFunction{InterRouteOptU}{InterRouteOptU}
    \SetKwFunction{GetSubTourU}{GetSubTourU}

    \Function{\InterRouteOptU{$r_1, r_2, p$}}{
        $r_1 \leftarrow$ \GetSubTourU{$r_1, p$}\;
        $r_2 \leftarrow$ \GetSubTourU{$r_2, p$}\;
         $\texttt{Improvement} \leftarrow \texttt{true}$\;
        \While{\texttt{Improvement} = \texttt{true}}{
            $\Delta_{best}, (a_1, a_2) \leftarrow$ \BestSwapInter{$r_1, r_2$}\;
            
            \If{$\Delta_{best} \geq 0$}{
                $\texttt{Improvement} \leftarrow \texttt{false}$\;
            }\uElse{
            \Swap{$r_1, r_2, a_1, a_2$}\;
            }
        }
        \Return{Optimized routes $r_1, r_2$}\;
    }
\end{algorithm}
\end{multicols}

\section{Hybrid algorithm}
\label{sec:methodology}

\begin{figure}
    \centering
    \includegraphics[width=\textwidth, trim={0cm 0.3cm 0cm 1.2cm}, clip]{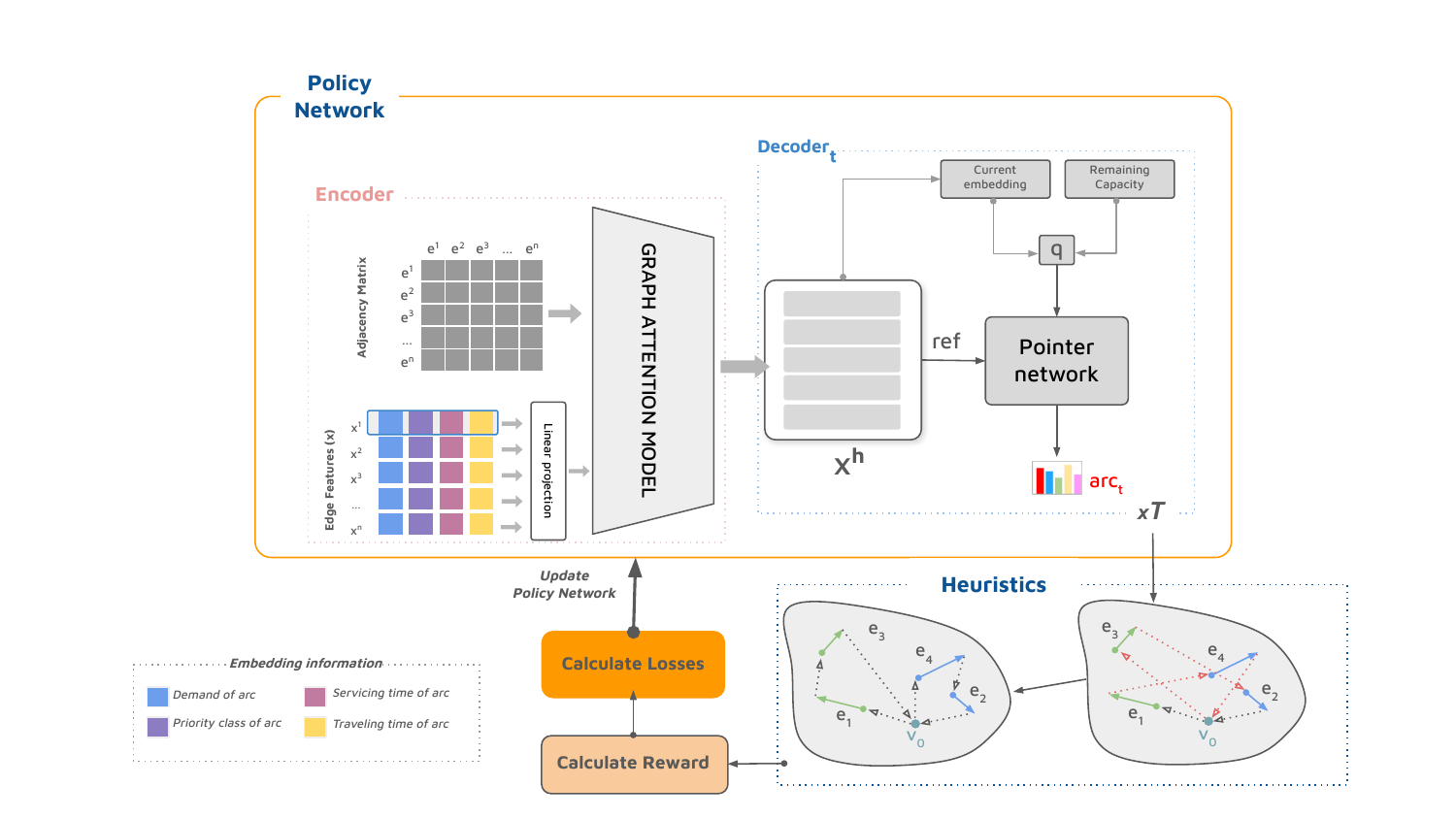}
    \caption{RL with Heuristic Integration.}
    \label{fig:overview}
\end{figure}

We introduce the hybrid HRDA algorithm that combines RL with the proposed heuristic approach. The solution is represented as a sequence of arcs, with each route separated by the depot, as shown below:
\[
\mathcal{R} = \{ v_0, r_1, v_0, r_2, \dots, v_0, r_m, v_0 \},
\]
where \( r_i \) represents a feasible route \((a_{i_1}, a_{i_2}, \dots, a_{i_k}) \), with \( v_0 \) as the depot and \( a_{i_1}, a_{i_2}, \dots, a_{i_k} \) as the arcs traversed sequentially within the route. The construction of these arc sequences is conceptualized as a Markov Decision Process (MDP). Initially, we outline the arcs properties and Adjacency matrix through a dedicated encoder model, named AMAE (Adjacency Matrix Attention Encoder) that is defined in Section \ref{method:amae}. After encoding the arcs into a latent space, they, together with the current state, are fed into a decoder, leveraging Autoregressive Sequence Generation based on a Pointer network\cite{bello2017neuralcombinatorialoptimizationreinforcement, vinyals2017pointernetworks}to iteratively produce the initial arc sequence.
 
Additionally, we employ intraroute and interroute swap operators as described in Section \ref{sec:Heuristic} to assist the policy network in selecting appropriate arcs, calculating rewards, and training the policy network within our hybrid algorithm. Figure \ref{fig:overview} illustrates all the steps involved in HRDA and details are given in Algorithm \ref{algo:HRDA}.

\subsection{Encoder}
\label{method:amae}

Inspired by the Transformer encoder \cite{kool2019attentionlearnsolverouting, jung2024lightweightcnntransformermodellearning}, which excel at learning arc properties, these methods, however, do not directly capture the distance information of arcs. Furthermore, the approach by \cite{kwon2021matrixencodingnetworksneural}  solely relies on adjacency matrix to learn distance information, omitting the arcs properties. To integrate the advantages and overcome the drawbacks of these previous studies, we introduce Adjacency Matrix Attention Encoder (AMAE), designed to generate embeddings directly from arcs, encapsulating both their attributes and distances—commonly associated with the adjacency matrix. Specifically, we handle $dx$-dimensional arc features that include distance to the depot, angle relative to the depot, demand, priority classes, servicing time, and traveling time of the arc. These features are subjected to a linear projection to produce initial embeddings $h_i^{(0)}$  and $\mathcal{D}$ (Adjacency Matrix), are processed through $N$ attention layers as follows:
$$
\hat{h}_i^{(l)} = \text{BN}^{(l)}\left(h_i^{(l-1)} + \text{MHA}_i^{(l)}\left(h_i^{(l-1)}, \mathcal{D}\right)\right),
$$
where $BN$ denotes Batch Normalization module, and MHA denotes Multi-Head Attention mechanism.
MHA, BN, and other useful neural architectures are mentioned in details in the Appendix \ref{appendix:GNN}.

\subsection{Decoder}
\label{method:decoder}


The output of the encoder is denoted by \( h_i^{(n)} = (h_1^{(n)}, h_2^{(n)}, \dots, h_m^{(n)}) \), where \( m \) represents the number of vehicles and \( n \) the layers of the encoder. The solution \( \mathcal{S} \) is subsequently decoded from \( h \) through a sequence of actions, formally expressed as:
\[
\alpha_t \sim \pi_{\theta}(\alpha_t \mid \alpha_{t-1}, \dots, \alpha_0, h),
\]
where \( \pi_\theta(\cdot) \) denotes the policy network and \( \theta \) represents the set of learnable parameters. The sequence of actions \( \boldsymbol{\alpha} = (\alpha_1, \dots, \alpha_m) \) collectively forms a feasible solution to the HDCARP.

\subsection{Training the policy network via reinforcement learning with heuristic integration}

\begin{figure}
    \centering
    \includegraphics[width=\textwidth, trim={0cm 0.3cm 0cm 1.2cm}, clip]{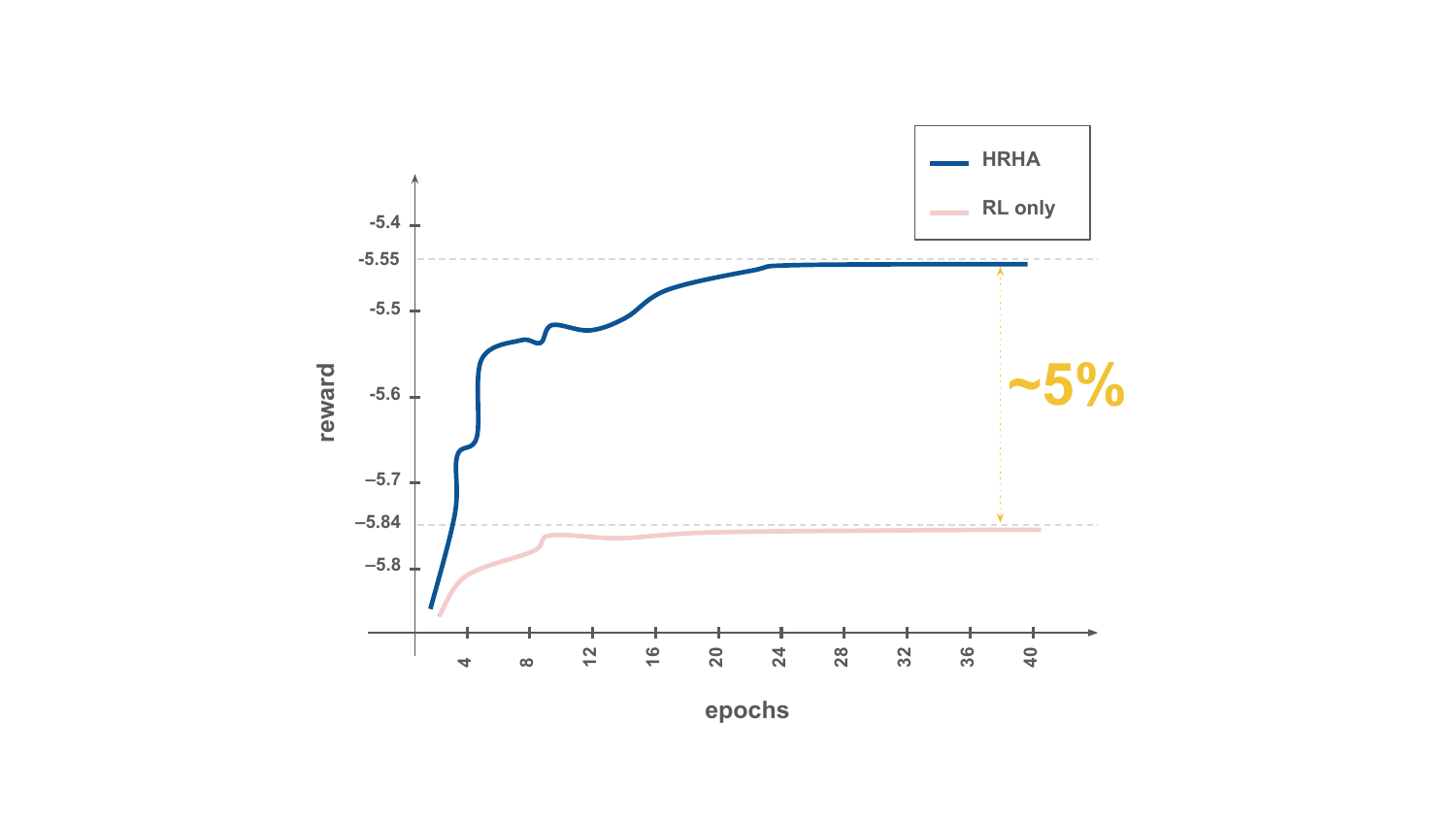}
    \caption{Comparison of HRDA and RL-only methods}
    \label{fig:compare_rlonly}
\end{figure}

The policy network \( \pi_\theta \), which includes both the encoder (see Section \ref{method:amae}) and the decoder (see Section \ref{method:decoder}), is trained using reinforcement learning methods, specifically the policy gradient method \cite{NIPS1999_464d828b}. This allows the model to learn from direct interactions with its environment. Initially, the RL model generates a solution at each decision point, which is then refined using a heuristic method. The reward is computed based on the improvement made by this heuristic enhancement. This hybrid approach, known as the hybrid HRDA algorithm, ensures that the RL agent not only learns effectively from environmental interactions but also benefits from the robust search capabilities of heuristic methods. The training objective is to optimize the policy network parameters \( \theta \) to maximize the expected reward, defined as the negative cost of the refined solution. The training objective is formally expressed as:
\[
\theta^* = \arg\max_\theta \mathbb{E}_{x \sim P(x)} \left[\mathbb{E}_{\boldsymbol{\alpha} \sim \pi_\theta(\boldsymbol{\alpha} \mid x)}[R(\boldsymbol{\alpha}, x)]\right],
\]
where \( P(x) \) represents the distribution of problem instances and \( R(\boldsymbol{\alpha}, x) \) is the reward function. The advantages of HRDA are clearly demonstrated in Figure \ref{fig:compare_rlonly}, showcasing its faster and more stable convergence and approximately $5\%$ improvement in reward performance compared to an RL-only approach. This strategic reduction allows the RL to focus more on optimizing the distribution of arcs across routes, rather than expending resources on exploring less impactful intraroute arc permutations.

\subsection{Reinforcement Learning and Heuristic Integration}
In this subsection, we describe the integration of Reinforcement Learning (RL) with heuristic methods to optimize solutions for Hierarchical Directed Capacitated Arc Routing Problems (HDCARP). By leveraging the adaptive capabilities of RL in conjunction with targeted heuristic operations, this hybrid method enhances both solution quality and computational efficiency.

\paragraph{Initial Solution Generation:} The RL agent starts by generating a preliminary sequence of routes, \(\mathcal{R} = \{ v_0, r_1, v_0, r_2, \dots, v_0, r_m, v_0 \}\), where each $r_k$ is a feasible route consisting of a sequence of arcs.

\paragraph{Intraroute Operation:} Each route $r_k$ undergoes optimization to improve its performance relative to the global objective. This is achieved by:
\[
r_k' = \arg\min_{r_k} \sum_{i=1}^{n_k} c(\alpha_i, \alpha_{i+1}),
\]
where $c(\alpha_i, \alpha_{i+1})$ is the cost function evaluating the expense between consecutive arcs $\alpha_i$ and $\alpha_{i+1}$ in route $r_k$. The focus here is on minimizing local costs by reordering arcs to enhance route efficiency.

\paragraph{Interroute Operation:} After internal optimizations, the next step involves redistributing arcs between routes to further minimize the maximum completion time, denoted by $C'_k$ for each route $k$. This step aims to adjust the routes such that the maximum of these completion times across all routes is minimized, reflecting the most time-critical route in the solution:
\[
a'' = \arg\min_{a'} \max_{k=1}^K C'_k,
\]
where $C'_k$ represents the total time to complete all tasks in route $k$ after redistribution. The objective is to ensure that the longest route is as short as possible, thereby optimizing the overall service efficiency and minimizing the maximum completion time under the constraint that:
\[
\sum_{e \in R'_k} d(e) \leq Q, \forall k,
\]
with $R'_k$ being the set of arcs in the newly optimized route $k$, and $d(e)$ representing the distance or cost associated with each arc $e$.

\paragraph{Reward Calculation and Policy Update:} The effectiveness of the newly optimized routes is quantified by calculating the reward $R(a'')$, which is defined as the negative of the maximum completion time $C''_{\text{max}}$:
\[
R(a'') = -C''_{\text{max}}.
\]
This reward is utilized to update the RL policy using Proximal Policy Optimization (PPO), focusing on balancing exploration and exploitation by adjusting the policy parameters based on the computed reward.

\begin{algorithm}
    \caption{Reinforcement Learning with Heuristic Optimization for HDCARP using PPO}
    \label{algo:HRDA}
    \SetKwInOut{Input}{Input}
    \SetKwInOut{Output}{Output}
    
    \Input{
        $\pi_{\theta}$: Initial policy network parameters;  $V_{\phi}$: Initial value network parameters;  \\
        $K$: Number of epochs; $T$: Number of episodes per epoch;\\ 
        $B$: Batch size; $\text{PPO\_epochs}$: Number of PPO updates per epoch;\\
    }
    \Output{Optimized policy parameters $\theta^*$ and value function parameters $\phi^*$}
    \BlankLine
    
    \SetKwFunction{LocalSearch}{LocalSearch}
    \SetKwFunction{SampleCARPInstance}{SampleCARPInstance}
    \SetKwFunction{GenerateInitialSolution}{GenerateInitialSolution}
    \SetKwFunction{PPOUpdate}{PPOUpdate}
    \SetKwFunction{HRDA}{HRDA}
    \SetKwFunction{FLSearch}{LocalSearch}
    \SetKwFunction{Random}{Random}
    \SetKwFunction{FIntra}{IntraSearch}
    \SetKwFunction{FInter}{InterSearch}
    \BlankLine
    
    \Function{\HRDA{$\pi_{\theta},\ V_{\phi},\ K,\ T,\ B$}}{
        \For{$k \leftarrow 1$ \KwTo $K$}{
            Initialize dataset $\mathcal{D} \leftarrow \emptyset$\;
            \For{$t \leftarrow 1$ \KwTo $T$}{
                $S \leftarrow$ \SampleCARPInstance{$B$}\;
                $A \leftarrow$ \GenerateInitialSolution{$S$, $\pi_{\theta}$}\;
                $A^* \leftarrow$ \LocalSearch{$A$}\;
                \ForEach{$(s_t, a_t^*, s_{t+1})$ in $(S, A^*, S')$}{
                    Compute reward $R_t$\;
                    \tcp{Store state, actions, rewards, negative log likelihood of actions}
                    Store $(s_t, a_t^*, R_t, nllg_{a_t})$ in $\mathcal{B}$\;
                }
            }
            Update $\theta$ and $\phi$ using \PPOUpdate{$\theta,\ \phi,\ \mathcal{B}$}\;
        }
        \Return{$\theta,\ \phi$}
    }

\end{algorithm}

\section{Metaheuristics}
\label{sec:meta}

Since the instances that can be optimally solved using the branch-and-cut algorithm from \cite{HDNNL24} are relatively small, we propose in this subsection three metaheuristics capable of producing high-quality solutions for larger instances to test with our hybrid algorithm. In particular, we designed an \emph{Iterated Local Search} (ILS), an \emph{Evolutionary Algorithm} (EA), and an \emph{Ant Colony Optimization} (ACO).

\subsection{Iterated Local Search (ILS)}
\label{method:ils}
Our ILS uses our construction phase to generate an initial solution, which is then improved by the local search phase described in Section \ref{sec:Heuristic}. To escape local optima and explore a wider search space, ILS applies a perturbation procedure to the best-known solution. Specifically, a priority class is randomly selected, along with a route that services at least two required arcs from this class. Two arbitrary arcs within this route are then chosen for swapping.  The perturbation and local search procedures are repeated iteratively for $k_{max}$ iterations.  After conducting preliminary experiments, we opted for $k_{max}= 10$. The new solution is accepted if an improvement is found. Details are given in Algorithm \ref{algo:ILS}.

\begin{algorithm}
    \caption{Iterated Local Search Algorithm}
    \label{algo:ILS}
    \SetKwInOut{Input}{Input}
    \SetKwInOut{Output}{Output}
    
    \Input{
        $k_{max}$: the number of iterations.
    }
    \Output{A solution $\mathcal{R}$}
    
    \SetKwFunction{ILS}{ILS}
    \BlankLine
    
    \Function{\ILS{$\mathcal{M}$}}{
        Call the \emph{greedy constructive heuristic} \ref{algo:NIH} to obtain a solution $\mathcal{R}$ \;
        $\mathcal{R}'$ = $\mathcal{R}$\;
        \For{$k = 1$ \KwTo $k_{max}$}{
            Perturb $\mathcal{R}'$ by randomly selecting a priority class and a route with at least two required arcs from that class, then swapping two arcs within the route\;
            Apply local search procedure in Subsection \ref{algo:localsearch} on $\mathcal{R}'$\;
            \If{$\mathcal{R}''$ better than {$\mathcal{R}$}}{
                $\mathcal{R}$ = $\mathcal{R}''$\;
                $\mathcal{R}'$ = $\mathcal{R}$\;
            }
        }
        \Return{$\mathcal{R}$}\;
    }
\end{algorithm}

\subsection{Evolutionary Algorithm (EA)}
\label{method:ea}

Our EA starts by generating an initial population $\Omega$ of $\lambda=200$ solutions using our greedy constructive heuristic \ref{algo:NIH}. In each iteration, the top four solutions are selected as parents. From these parents, segments within a randomly chosen priority class are swapped to create offspring. If capacity constraints are violated, smaller sub-segments are swapped instead. Each offspring undergoes the local search procedure described in Subsection \ref{sec:Heuristic} before being added back to the population. After $k_{\text{max}} = 100$ iterations, the algorithm stops and returns the best solution $\mathcal{R}$ found. Details are
given in Algorithm \ref{algo:evolutionary-algorithm}.

\begin{algorithm}
    \caption{Evolutionary Algorithm}
    \label{algo:evolutionary-algorithm}
    \SetKwInOut{Input}{Input}
    \SetKwInOut{Output}{Output}
    
    \Input{
        $\lambda$: population size;\\
        $k_{max}$: the number of iterations.
    }
    \Output{A solution $\mathcal{R}$}

    \SetKwFunction{EA}{EA}
    \BlankLine
    
    \Function{\EA{$k_{max}, \lambda$}}{
        Use the \emph{greedy constructive heuristic} \ref{algo:NIH}
        to generate an initial set of solutions \(\Omega\) of size \(\lambda\)  \;
        \For{$k = 1$ \KwTo $k_{max}$}{
            \Indp
            Select the top 4 solutions from \(\Omega\) as parents\;

            Randomly select a priority class $p$ \;
            
            Extract and swap segments of class $p$ from the parents to create offspring\;

            \If{capacity constraints are violated}{
            Adjust offspring by swapping smaller sub-segments from the parents\;
            }
            
            Apply the local search procedure in Subsection \ref{algo:localsearch} to all offspring and include them in \(\Omega\)\;
        }

        \BlankLine
        Let \(\mathcal{R} \) be the best solution in \(\Omega\) \;
        \Return{\(\mathcal{R}\)}
    }
\end{algorithm}

\subsection{Ant Colony Optimization (ACO)}
\label{method:aco}
Our ACO starts by creating a matrix of size $|A_r| \times |A_r|$ to store pheromone levels. Initially, all pheromone values are set to zero. The algorithm uses 50 ants. In each iteration, each ant builds a feasible solution by choosing arcs according to pheromone levels. After an ant creates a solution, the local search procedure, described in Subsection \ref{sec:Heuristic}, is applied to improve that solution. The pheromone values are updated, reinforcing the arcs used by that solution and evaporating all pheromone values by a factor of 0.5. These steps are repeated for 100 iterations, after which the best solution $\mathcal{R}$ is returned. Details are
given in Algorithm \ref{algo:ant-colony-algorithm}.

\begin{algorithm}[H]
    \caption{Ant Colony Optimization}
    \label{algo:ant-colony-algorithm}
    \SetKwInOut{Input}{Input}
    \SetKwInOut{Output}{Output}

    \Input{
        $n_{ant}$: number of ants; $k_{max}$: maximum number of iterations;
        $\rho=0.5$
    }
    \Output{A solution $\mathcal{R}$}

    \SetKwFunction{ACO}{ACO}
    \BlankLine
    
    \Function{\ACO{$n$, $n_{ant}, k_{max}$}}{
        Initialise a pheromone matrix of size $|A_r| \times |A_r|$ with zero entries\\
        \For{$k = 1$ \KwTo $k_{max}$}{
            Let $\Omega$ be an empty set of solutions\;
            \For{each of the $n_{ant}$}{
                Construct a feasible solution using pheromones\\
                Apply the local search procedure in Subsection \ref{algo:localsearch} to improve the solution and add it to $\Omega$\\
                
            }
            Increase pheromones along arcs used in the solution obtained\;
            Evaporate all pheromones by a factor of $(1-\rho)$\;
        }
        Let \(\mathcal{R} \) be the best solution \;
        \Return{\(\mathcal{R}\)}
    }
\end{algorithm}

\section{Computational experiments}
\label{sec:Experimental}
In this section, we present the results of some computational experiments. For all of our experiments, we used a PC with a 13th Gen Intel(R) Core(TM) i7-13700K processor, running Ubuntu with 64GB of RAM. All algorithms were implemented in Python, and the LP relaxations were solved using the Gurobi solver.

\subsection{Create HDCARP instances}

We create HDCARP instances as follows:

\begin{enumerate}
    \item Graph extraction: Use the Python package {\tt OSMnx} \cite{Bo17} to extract a strongly connected graphs \( G = (V, A) \)  with a specific number of arcs $|A| \in \{30, 40, \dots, 200\}$ from {\tt OpenStreetMap}.
    \item Vehicle fleet: The number of vehicles is chosen from the set ${2, 5}$. For each combination of \( |M| \) and \( |A| \), we generate a total of 20 instances. Specifically, for \( |M| = 2 \), the values of \( |A| \) range from 30 to 200, resulting in 360 instances. For \( |M| = 5 \), the values of \( |A| \) range from 70 to 200, giving a total of 280 instances.
    \item Set of required arcs: Divide the arc set \( A \) into \( A_r \) and \( A_{nr} \). If \( |A| \ge 80 \), \( |A_r| \) is randomly chosen between 60 and 70; if \( |A| < 80 \), \( |A_r| \) is set to $75\%$ of \( |A| \). 
    \item Priority classes: Let \( P = \{1, 2, 3\} \) and the class of each required arc is randomly assigned from 1 to 3. 
    \item Traversal and servicing time: Compute the Euclidean distance $d'_a$ for each arc $a$, and then normalise these distances to obtain the traversal time as $d_a = \frac{d'_a}{d'_{\text{max}}}$, where $d'_{\text{max}}$ denotes the maximum Euclidean distance among all arcs. The service time for each required arc is set to be twice the traversal time.
    \item Demand of required arcs: Calculate the demand for each arc \( a \in A_r \) as \( q_a = d_a \times 0.5 + 0.5 \).
    \item Vehicle capacity: Calculate the vehicle capacity $Q$ as \( Q = \sum_{a \in A_r} \left(\frac{q_a}{3} + 0.5\right) \).
\end{enumerate}

Comprehensive details of all instances, along with the source codes, will be made available on GitHub at \href{https://github.com/HySonLab/ArcRoute}{https://github.com/HySonLab/ArcRoute}.

We additionally construct a training dataset of 1 million instances to train the model. All instances in this dataset share the same parameters: $|A_r| = 100$, $p = 3$, and $|M| = 5$. While the training set follows a fixed structure, our algorithms are reliably capable of handling instances with varying numbers of arcs, as will be confirmed by experimental results.

\subsection{Computational results}

To test the performance of our hybrid algorithm HRDA, we compared it with our heuristics described in Section \ref{sec:Heuristic}, the \emph{Evolutionary Algorithm} (EA) described in Appendix \ref{method:ea}, the \emph{Ant Colony Optimization} (ACO) described in Appendix \ref{method:aco}, and the \emph{Exact Method} (EM) from \cite{HDNNL24}, described in Appendix \ref{method:exact}.

The results for HDCARP-P variant that we obtained on the group of artificial instances with $|M|=2$ are shown in Table \ref{tab:2mp}. The first column shows the number of arcs $|A|$. As mentioned in Section \ref{sec:intro}, a solution is considered better if it reduces the time taken to complete tasks in higher-priority classes. All algorithms tested for this variant produce a solution with a non-zero gap in the maximum completion time of the first priority class. Therefore, for the EM algorithm, we report the average completion time of the first priority class (``Obj''), whereas for other algorithms, we report the average optimality gap of the completion time for the first priority class (``Gap''), and the average overall running time (``Time''), in seconds, across all $20$ instances. We highlight the best gap of each algorithm (except for the EM) in red, and the second-best gap in orange.

In terms of solution quality, EA generally performs closest to the EM, followed by HRDA. However, when $|A| \ge 160$, HRDA outperforms EA. HRDA also stands out for producing solutions with consistent quality, as indicated by the smallest gap deviation. In terms of running time, although ILS is the fastest, its solution quality is the poorest. HRDA follows closely behind. ACO and EA are the slowest among the four algorithms.

\begin{table}[htbp]
\centering
\caption{Comparison of all algorithms for the HDCARP-P variant on instances with 2 vehicles. Red color denotes the best performance while orange color denotes the second-best one.}
\label{tab:2mp}
\begin{tabular}{@{}ccccccccccccccc@{}}
\toprule
\multirow{2}{*}{$|A|$} & \multicolumn{2}{c}{EM} & \multicolumn{2}{c}{ILS} & \multicolumn{2}{c}{ACO} & \multicolumn{2}{c}{EA} & \multicolumn{2}{c}{HRDA} (Ours) \\
\cmidrule(r){2-3} \cmidrule(l){4-5} \cmidrule(l){6-7} \cmidrule(l){8-9} \cmidrule(l){10-11}
 & obj & time & gap & time & gap & time & gap & time &  gap & time \\
\midrule
30 & 6.66 & 0.51 & 6.03 & 0.01 & \textcolor{orange}{1.14} & 197.94 & \textcolor{red}{1.03} & 38.21 & 1.38 & 0.42 \\
40 & 6.9 & 0.82 & 6.45 & 0.02 & 0.88 & 104.96 & \textcolor{red}{0.54} & 40.17 & \textcolor{orange}{0.80} & 0.41 \\
50 & 7.01 & 1.3 & 3.83 & 0.02 & 1.68 & 212.42 & \textcolor{red}{0.70} & 42.66 & \textcolor{orange}{0.98} & 0.42 \\
60 & 7.59 & 2.37 & 7.34 & 0.02 & 1.24 & 113.84 & \textcolor{red}{0.60} & 45.17 & \textcolor{orange}{0.96} & 0.41 \\
70 & 8.25 & 4.28 & 4.26 & 0.02 & \textcolor{orange}{1.09} & 176.6 & \textcolor{red}{0.95} & 47.79 & \textcolor{red}{1.13} & 0.41 \\
80 & 7.68 & 5.44 & 4.45 & 0.01 & \textcolor{orange}{1.21} & 97.08 & \textcolor{red}{1.19} & 50.61 & 1.34 & 0.42 \\
90 & 8.22 & 11.11 & 1.75 & 0.02 & \textcolor{red}{0.68} & 174.43 & \textcolor{orange}{0.68} & 52.5 & 0.92 & 0.43 \\
100 & 8.65 & 20.72 & 4.19 & 0.02 & 0.99 & 189.11 & \textcolor{red}{0.89} & 171.92 & \textcolor{orange}{0.95} & 0.4 \\
110 & 8.18 & 246.56 & 4.53 & 0.02 & 1.08 & 152.12 & \textcolor{red}{0.74} & 123.79 & \textcolor{orange}{1.06} & 0.43 \\
120 & 8.34 & 270.49 & 6.05 & 0.02 & \textcolor{orange}{0.90} & 243.8 & \textcolor{red}{1.06} & 278.83 & 1.22 & 0.42 \\
130 & 9.12 & 70.77 & 5.57 & 0.02 & 1.38 & 234.07 & \textcolor{red}{0.79} & 145.44 & \textcolor{orange}{0.97} & 0.41 \\
140 & 10.52 & 198.92 & 5.78 & 0.02 & 1.47 & 132.1 & \textcolor{red}{0.95} & 169.01 & \textcolor{orange}{1.21} & 0.41 \\
150 & 9.07 & 199.52 & 5.45 & 0.02 & 0.97 & 123.88 & \textcolor{red}{0.88} & 173.46 & \textcolor{orange}{0.91} & 0.42 \\
160 & 10.3 & 242.59 & 5.27 & 0.02 & 1.44 & 177.72 & \textcolor{orange}{1.20} & 177.98 & \textcolor{red}{1.13} & 0.44 \\
170 & 9.71 & 282.59 & 5.68 & 0.02 & 2.34 & 123.37 & \textcolor{orange}{1.31} & 182.28 & \textcolor{red}{1.27} & 0.44 \\
180 & 11.61 & 243.68 & 5.55 & 0.02 & 1.21 & 210.63 & \textcolor{orange}{0.94} & 186.98 & \textcolor{red}{0.83} & 0.42 \\
190 & 11.36 & 399 & 5.78 & 0.02 & 1.05 & 364.29 & \textcolor{red}{0.96} & 192.38 & \textcolor{orange}{1.04} & 0.42 \\
200 & 10.7 & 440 & 5.7 & 0.02 & 1.39 & 323.86 & \textcolor{red}{1.26} & 298.21 & \textcolor{red}{1.05} & 0.43 \\
\bottomrule
\end{tabular}
\end{table}

Table \ref{tab:2mu} shows the results for the HDCARP-U variant on the same set of instances and  has an identical format to Table \ref{tab:2mp}. While all algorithms produce the same maximum completion time for the first priority class, the results differ for the second priority class. Therefore, we report the values for the second priority class. 

The results of ILS here are similar to what we saw in the case of the HDCARP-P variant. Therefore, we exclude the ILS from our consideration. In terms of solution quality, ACO outperforms both EA and HRDA overall. However, it is the slowest of the three algorithms, with a running time comparable to that of EM. HRDA is comparable to EA, but performs better on larger-scale instances, and is the fastest among the three. 

These results demonstrates that our hybrid algorithm is capable of providing solutions of very good quality within a short computing time, making it suitable for large-scale instances.

\begin{table}[htbp]
\centering
\caption{Comparison of all algorithms for the HDCARP-U variant on instances with 2 vehicles. Red color denotes the best performance while orange color denotes the second-best one.}
\label{tab:2mu}
\begin{tabular}{@{}ccccccccccccccc@{}}
\toprule
\multirow{2}{*}{$|A|$} & \multicolumn{2}{c}{EM} & \multicolumn{2}{c}{ILS} & \multicolumn{2}{c}{ACO} & \multicolumn{2}{c}{EA} & \multicolumn{2}{c}{HRDA} \\
\cmidrule(r){2-3} \cmidrule(l){4-5} \cmidrule(l){6-7} \cmidrule(l){8-9} \cmidrule(l){10-11}
 & obj & time & gap & time & gap & time & gap & time &  gap & time \\
\midrule
30 & 12.45 & 24.43 & 3.58 & 0.02 & \textcolor{orange}{1.87} & 19.13 & \textcolor{red}{1.69} & 66.68 & 2.48 & 0.64 \\
40 & 13.95 & 2.23 & 4.65 & 0.02 & \textcolor{red}{0.67} & 25.2 & \textcolor{orange}{0.71} & 71 & 0.98 & 0.59 \\
50 & 13.32 & 3.52 & 5.85 & 0.02 & \textcolor{red}{1.50} & 34.21 & 2.01 & 75.32 & \textcolor{orange}{1.84} & 0.55 \\
60 & 15.25 & 6.46 & 10.68 & 0.02 & \textcolor{red}{1.89} & 45.19 & \textcolor{orange}{2.16} & 78.68 & 2.31 & 0.51 \\
70 & 15.62 & 11.64 & 7.55 & 0.02 & \textcolor{red}{2.12} & 60.64 & \textcolor{orange}{2.40} & 102.56 & 2.7 & 0.57 \\
80 & 15.67 & 14.79 & 7.53 & 0.02 & \textcolor{red}{2.56} & 81.14 & \textcolor{orange}{2.76} & 102.7 & 2.81 & 0.6 \\
90 & 16.33 & 30.2 & 7.51 & 0.02 & \textcolor{orange}{2.76} & 108.26 & \textcolor{red}{2.70} & 131.28 & 2.95 & 0.66 \\
100 & 16.26 & 56.34 & 9.27 & 0.03 & \textcolor{red}{2.22} & 147.83 & 3.05 & 147.83 & \textcolor{orange}{2.44} & 0.54 \\
110 & 17.1 & 670.21 & 8.65 & 0.03 & \textcolor{red}{2.86} & 189.52 & 3.31 & 159.74 & \textcolor{orange}{3.12} & 0.63 \\
120 & 16.17 & 735.26 & 11.71 & 0.03 & \textcolor{red}{4.82} & 167.38 & 6.03 & 131.52 & \textcolor{orange}{5.19} & 0.6 \\
130 & 18.15 & 192.38 & 8.36 & 0.03 & \textcolor{red}{3.73} & 158.8 & 4.06 & 117.54 & \textcolor{orange}{3.80} & 0.53 \\
140 & 18.46 & 140.73 & 9.68 & 0.03 & \textcolor{orange}{3.68} & 169.06 & \textcolor{red}{3.55} & 123.1 & 3.77 & 0.56 \\
150 & 17.47 & 242.36 & 8.92 & 0.03 & \textcolor{orange}{3.22} & 243.36 & 3.82 & 130.63 & \textcolor{red}{3.15} & 0.65 \\
160 & 20.08 & 221.36 & 8.13 & 0.03 & \textcolor{orange}{2.46} & 254.65 & \textcolor{red}{2.27} & 137.49 & 2.78 & 0.64 \\
170 & 18.88 & 294.65 & 9.73 & 0.03 & \textcolor{orange}{4.40} & 255.23 & 4.42 & 145.03 & \textcolor{red}{4.27} & 0.66 \\
180 & 21.14 & 289 & 6.25 & 0.03 & \textcolor{red}{4.17} & 221.14 & 4.24 & 153.12 & \textcolor{orange}{4.21} & 0.49 \\
190 & 21.6 & 347.34 & 5.92 & 0.02 & \textcolor{red}{1.18} & 285.26 & \textcolor{orange}{1.45} & 161.86 & 1.48 & 0.57 \\
200 & 20.68 & 367.89 & 7.11 & 0.03 & \textcolor{red}{1.17} & 312.76 & 2.29 & 169.56 & \textcolor{orange}{1.29} & 0.58 \\
\bottomrule
\end{tabular}
\end{table}

\begin{table}[htbp]
\centering
\caption{Comparison between the two HDCARP variants on instances with 2 vehicles. Bold text denotes better performance between P and U variants. In our observation, U variant always outperforms the P variant.}
\label{tab:2UP}
\begin{tabular}{@{}ccccccccccc@{}}
\toprule
\multirow{2}{*}{$|A|$} & \multicolumn{2}{c}{EM} & \multicolumn{2}{c}{ILS} & \multicolumn{2}{c}{ACO} & \multicolumn{2}{c}{EA} & \multicolumn{2}{c}{HRDA} \\
\cmidrule(r){2-3} \cmidrule(l){4-5} \cmidrule(l){6-7} \cmidrule(l){8-9} \cmidrule(l){10-11}
 & P  & U & P  & U & P  & U & P  & U & P  & U \\
\midrule
30 & 12.94 & \textbf{12.45} & 16.49 & \textbf{12.87} & 14.72 & \textbf{12.67} & 15.03 & \textbf{12.64} & 15.28 & \textbf{12.74} \\
40 & 14.13 & \textbf{13.95} & 18.08 & \textbf{14.61} & 15.31 & \textbf{14.04} & 16.48 & \textbf{14.05} & 16.75 & \textbf{14.08} \\
50 & 13.44 & \textbf{13.32} & 17.45 & \textbf{14.10} & 15.67 & \textbf{13.51} & 15.91 & \textbf{13.58} & 16.17 & \textbf{13.56} \\
60 & 15.26 & \textbf{15.25} & 20.51 & \textbf{16.86} & 17.61 & \textbf{15.53} & 18.7 & \textbf{15.58} & 19.01 & \textbf{15.60} \\
70 & 15.75 & \textbf{15.62} & 19.92 & \textbf{16.78} & 17.13 & \textbf{15.93} & 18.16 & \textbf{15.99} & 18.46 & \textbf{16.03} \\
80 & 15.89 & \textbf{15.67} & 21.34 & \textbf{16.84} & 19.19 & \textbf{16.06} & 19.45 & \textbf{16.09} & 19.77 & \textbf{16.10} \\
90 & 16.36 & \textbf{16.33} & 21.54 & \textbf{17.55} & 19.87 & \textbf{16.76} & 19.64 & \textbf{16.75} & 19.96 & \textbf{16.79} \\
100 & 16.41 & \textbf{16.26} & 21.68 & \textbf{17.75} & 19.98 & \textbf{16.63} & 19.76 & \textbf{16.76} & 20.08 & \textbf{16.67} \\
110 & 17.16 & \textbf{17.10} & 21.28 & \textbf{18.61} & 18.76 & \textbf{17.58} & 19.4 & \textbf{17.68} & 19.72 & \textbf{17.62} \\
120 & 16.23 & \textbf{16.17} & 20.29 & \textbf{17.97} & 18.19 & \textbf{16.89} & 18.5 & \textbf{17.07} & 18.8 & \textbf{16.95} \\
130 & 18.18 & \textbf{18.15} & 23.62 & \textbf{19.67} & 20.07 & \textbf{18.82} & 21.53 & \textbf{18.89} & 21.88 & \textbf{18.84} \\
140 & 18.51 & \textbf{18.46} & 25.11 & \textbf{20.22} & 21.49 & \textbf{19.14} & 22.89 & \textbf{19.12} & 23.27 & \textbf{19.15} \\
150 & 17.56 & \textbf{17.47} & 24.19 & \textbf{19.09} & 21.34 & \textbf{18.03} & 22.05 & \textbf{18.14} & 22.41 & \textbf{18.01} \\
160 & 20.6 & \textbf{20.08} & 26.59 & \textbf{21.68} & 24.87 & \textbf{20.51} & 24.24 & \textbf{20.49} & 24.64 & \textbf{20.58} \\
170 & 19.33 & \textbf{18.88} & 24.91 & \textbf{20.63} & 22.89 & \textbf{19.60} & 22.71 & \textbf{19.59} & 23.08 & \textbf{19.57} \\
180 & 21.16 & \textbf{21.14} & 27.86 & \textbf{22.42} & 24.16 & \textbf{21.94} & 25.4 & \textbf{21.95} & 25.82 & \textbf{21.94} \\
190 & 22.11 & \textbf{21.60} & 27.67 & \textbf{22.90} & 24.13 & \textbf{21.87} & 25.16 & \textbf{21.90} & 25.57 & \textbf{21.92} \\
200 & 21.09 & \textbf{20.68} & 26.34 & \textbf{22.13} & 24.33 & \textbf{20.93} & 24.01 & \textbf{21.14} & 24.4 & \textbf{20.95} \\
\bottomrule
\end{tabular}
\end{table}

\begin{table}[htbp]
\centering
\caption{Comparison of all algorithms for the HDCARP-P variant on instances with 5 vehicles. Red color denotes the best performance while orange color denotes the second-best one.}
\label{tab:3mp}
\begin{tabular}{@{}ccccccccc@{}}
\toprule
\multirow{2}{*}{$|A|$} & \multicolumn{2}{c}{ILS} & \multicolumn{2}{c}{ACO} & \multicolumn{2}{c}{EA} & \multicolumn{2}{c}{HRDA} \\
\cmidrule(r){2-3} \cmidrule(l){4-5} \cmidrule(l){6-7} \cmidrule(l){8-9} 
 & obj  & time & obj  & time & obj  & time & obj  & time \\
\midrule
70 & 9.80 & 0.06 & \textcolor{red}{8.39} & 430.93 & 9.05 & 136.87 & \textcolor{orange}{8.54} & 1.24 \\
80 & 10.17 & 0.06 & \textcolor{red}{8.71} & 437.92 & 9.53 & 151.51 & \textcolor{orange}{8.90} & 1.52 \\
90 & 10.35 & 0.06 & \textcolor{red}{8.91} & 423.67 & 9.67 & 169.88 & \textcolor{orange}{9.07} & 0.56 \\
100 & 11.22 & 0.06 & \textcolor{red}{9.35} & 413.33 & 10.09 & 247.94 & \textcolor{orange}{9.52} & 1.06 \\
110 & 10.77 & 0.06 & \textcolor{red}{9.17} & 431.02 & 9.85 & 157.24 & \textcolor{orange}{9.30} & 1.37 \\
120 & 11.25 & 0.05 & \textcolor{red}{9.59} & 428.52 & 10.16 & 177.28 & \textcolor{orange}{9.86} & 1.28 \\
130 & 12.38 & 0.05 & \textcolor{red}{10.21} & 413.86 & 11.05 & 193.22 & \textcolor{orange}{10.41} & 1.16 \\
140 & 12.08 & 0.06 & \textcolor{red}{10.07} & 415.20 & 10.79 & 213.22 & \textcolor{orange}{10.31} & 1.35 \\
150 & 12.32 & 0.06 & \textcolor{red}{10.33} & 416.01 & 11.21 & 216.65 & \textcolor{orange}{10.51} & 1.03 \\
160 & 13.78 & 0.06 & \textcolor{red}{11.55} & 442.55 & 12.44 & 249.91 & \textcolor{orange}{11.79} & 1.43 \\
170 & 12.60 & 0.06 & \textcolor{red}{10.51} & 422.75 & 11.35 & 267.09 & \textcolor{orange}{10.72} & 1.11 \\
180 & 14.96 & 0.06 & \textcolor{red}{12.54} & 481.43 & 13.60 & 286.46 & \textcolor{orange}{12.85} & 2.04 \\
190 & 14.06 & 0.06 & \textcolor{red}{11.89} & 445.20 & 12.82 & 267.96 & \textcolor{orange}{12.19} & 2.13 \\
200 & 14.58 & 0.06 & \textcolor{red}{12.12} & 434.47 & 13.07 & 337.93 & \textcolor{orange}{12.45} & 2.39 \\
\bottomrule
\end{tabular}
\end{table}

\begin{table}[htbp]
\centering
\caption{Comparison of all algorithms for the HDCARP-U variant on instances with 5 vehicles. Red color denotes the best performance while orange color denotes the second-best one.}
\label{tab:3mu}
\begin{tabular}{@{}ccccccccc@{}}
\toprule
\multirow{2}{*}{$|A|$} & \multicolumn{2}{c}{ILS} & \multicolumn{2}{c}{ACO} & \multicolumn{2}{c}{EA} & \multicolumn{2}{c}{HRDA} \\
\cmidrule(r){2-3} \cmidrule(l){4-5} \cmidrule(l){6-7} \cmidrule(l){8-9} 
 & obj  & time & obj  & time & obj  & time & obj  & time \\
\midrule
70 & 18.22 & 0.06 & \textcolor{red}{16.43} & 430.93 & 17.78 & 136.87 & \textcolor{orange}{17.87} & 1.40 \\
80 & 19.90 & 0.06 & \textcolor{red}{17.63} & 437.92 & 19.79 & 151.51 & \textcolor{orange}{19.19} & 1.56 \\
90 & 19.97 & 0.06 & \textcolor{red}{17.89} & 423.67 & 19.99 & 169.88 & \textcolor{orange}{18.63} & 0.86 \\
100 & 21.17 & 0.06 & \textcolor{red}{18.82} & 413.33 & \textcolor{orange}{20.01} & 247.94 & 20.14 & 1.12 \\
110 & 20.80 & 0.06 & \textcolor{red}{18.30} & 431.02 & 19.94 & 157.24 & \textcolor{orange}{19.04} & 1.60 \\
120 & 21.68 & 0.06 & \textcolor{red}{18.59} & 428.52 & 20.40 & 177.28 & \textcolor{orange}{19.28} & 1.85 \\
130 & 23.36 & 0.06 & \textcolor{red}{20.39} & 413.86 & 22.23 & 193.22 & \textcolor{orange}{21.97} & 1.39 \\
140 & 22.92 & 0.06 & \textcolor{red}{20.07} & 415.20 & 22.21 & 213.22 & \textcolor{orange}{21.67} & 1.77 \\
150 & 23.50 & 0.06 & \textcolor{red}{20.35} & 416.01 & 22.04 & 216.65 & \textcolor{orange}{21.85} & 1.33 \\
160 & 25.13 & 0.06 & \textcolor{red}{21.45} & 442.55 & 23.27 & 249.91 & \textcolor{orange}{23.74} & 1.96 \\
170 & 24.38 & 0.06 & \textcolor{red}{21.01} & 422.75 & 22.83 & 267.09 & \textcolor{orange}{22.29} & 1.21 \\
180 & 26.89 & 0.06 & \textcolor{orange}{23.68} & 481.43 & 24.28 & 286.46 & \textcolor{red}{23.05} & 2.24 \\
190 & 26.71 & 0.06 & \textcolor{red}{23.03} & 445.20 & 24.80 & 267.96 & \textcolor{orange}{23.79} & 2.44 \\
200 & 27.37 & 0.06 & \textcolor{red}{23.35} & 434.47 & 25.11 & 337.93 & \textcolor{orange}{24.97} & 1.93 \\
\bottomrule
\end{tabular}
\end{table}

\begin{table}[htbp]
\centering
\caption{Comparison between the two HDCARP variants on instances with 5 vehicles. Bold text denotes better performance between P and U variants. In our observation, U variant always outperforms the P variant.}
\label{tab:5UP}
\begin{tabular}{@{}ccccccccc@{}}
\toprule
\multirow{2}{*}{$|A|$} & \multicolumn{2}{c}{ILS} & \multicolumn{2}{c}{ACO} & \multicolumn{2}{c}{EA} & \multicolumn{2}{c}{HRDA} \\
\cmidrule(r){2-3} \cmidrule(l){4-5} \cmidrule(l){6-7} \cmidrule(l){8-9}
 & P  & U & P  & U & P  & U & P  & U \\
\midrule
70 & 18.76 & \textbf{18.22} & 17.15 & \textbf{16.43} & 18.68 & \textbf{17.78} & 18.06 & \textbf{17.87} \\
80 & 20.64 & \textbf{19.90} & 18.00 & \textbf{17.63} & 20.59 & \textbf{19.79} & 19.60 & \textbf{19.19} \\
90 & 20.25 & \textbf{19.97} & 18.11 & \textbf{17.89} & 20.06 & \textbf{19.99} & 19.24 & \textbf{18.63} \\
100 & 21.91 & \textbf{21.17} & 19.69 & \textbf{18.82} & 20.85 & \textbf{20.01} & 21.08 & \textbf{20.14} \\
110 & 21.15 & \textbf{20.80} & 19.06 & \textbf{18.30} & 20.41 & \textbf{19.94} & 19.83 & \textbf{19.04} \\
120 & 22.54 & \textbf{21.68} & 18.71 & \textbf{18.59} & 21.22 & \textbf{20.40} & 20.02 & \textbf{19.28} \\
130 & 24.24 & \textbf{23.36} & 21.04 & \textbf{20.39} & 22.68 & \textbf{22.23} & 22.23 & \textbf{21.97} \\
140 & 23.83 & \textbf{22.92} & 20.51 & \textbf{20.07} & 22.40 & \textbf{22.21} & 22.22 & \textbf{21.67} \\
150 & 23.60 & \textbf{23.50} & 20.44 & \textbf{20.35} & 22.17 & \textbf{22.04} & 22.50 & \textbf{21.85} \\
160 & 25.17 & \textbf{25.13} & 22.25 & \textbf{21.45} & 23.68 & \textbf{23.27} & 24.27 & \textbf{23.74} \\
170 & 24.43 & \textbf{24.38} & 21.26 & \textbf{21.01} & 23.13 & \textbf{22.83} & 22.62 & \textbf{22.29} \\
180 & 27.13 & \textbf{26.89} & 24.16 & \textbf{23.68} & 24.40 & \textbf{24.28} & 23.27 & \textbf{23.05} \\
190 & 27.09 & \textbf{26.71} & 23.37 & \textbf{23.03} & 25.10 & \textbf{24.80} & 24.43 & \textbf{23.79} \\
200 & 27.43 & \textbf{27.37} & 23.75 & \textbf{23.35} & 25.23 & \textbf{25.11} & 24.98 & \textbf{24.97} \\
\bottomrule
\end{tabular}
\end{table}

For interest, Table \ref{tab:2UP} \ref{tab:5UP} compares the results between the HDCARP variants in terms of the maximum completion time of the second priority class. It is important to note that, due to the nature of the hierarchical objective function, the optimal solution for both variants must have the same completion time for the first priority class. Additionally, as previously mentioned, all algorithms in the HDCARP-U variant produce solutions with the same completion time for the first priority class as the EM. So, for a detailed comparison of the two variants with respect to the maximum completion time of the first priority class, we refer readers to Table \ref{tab:2mp}.

One can see that the HDCARP-U variant significantly improves the service of the second priority class compared to the HDCARP-P variant for almost all instances. This improvement is undoubtedly due to the flexibility of servicing some required arcs of lower-priority classes before those of higher-priority classes, thereby reducing the time spent traveling without servicing. Compared to the EM, the optimality gap in the completion time of the second priority class for the HDCARP-U variant is considerably smaller than that of the HDCARP-P variant. This suggests that the swap operators are effective at escaping local optima for the HDCARP-U variant, but the HDCARP-P variant.

Tables \ref{tab:3mp} and \ref{tab:3mu} show the results for the HDCARP-P and HDCARP-U variants on the second group of instances with 5 vehicles. These tables have an identical format to Tables \ref{tab:2mp} and \ref{tab:2mu}, respectively, except that these instances are too large to solve to optimality by the EM. Among the four algorithms, ACO provides the best quality solutions, although it has the longest running time. In contrast, ILS is the fastest but produces the lowest quality solutions. HRDA's solution quality is comparable to that of ACO, while its running time consistently remains under 3 seconds, making it the best compromise between solution quality and running time, and thus suitable for large-scale instances.

\section{Conclusion}
\label{sec:conclusion}
In this study, we proposed our hybrid algorithm that integrates RL with a heuristic approach to efficiently solve HDCARP variants. By exploiting the adaptive learning capabilities of RL and the efficiency of heuristics in producing high-quality solutions within a short amount of time, our approach addresses key challenges related to stability, convergence, and computational efficiency that are commonly encountered in standalone RL techniques. The hybrid algorithm effectively guides local search operators, speeds up convergence, and improves the solution quality. Extensive computational experiments have demonstrated the effectiveness of our hybrid algorithm compared to traditional heuristics, metaheuristics, and standalone RL approaches, making it particularly suitable for large-scale instances.

We can think of two possible topics for future research. The first is the development of local search operators to improve the solution for HDCARP-P variant. The second is the adaptation of our hybrid algorithm to other ARPs, such as ARPs with time deadlines \cite{Li92, EL96}, time windows \cite{Eg94,LS87}, and/or multiple depots \cite{Eg94}. The third is the integration of multi-agent reinforcement learning to improve collaborative routing strategies across fleets.





\bibliographystyle{elsarticle-harv}
\bibliography{main}






\section{Appendix}
\label{sec:appendix}
\subsection{Exact Method (EM)}
\label{method:exact}

We present the branch-and-cut algorithms to solve both HDCARP variants, as described in \cite{HDNNL24}. 

\subsubsection{Graph transformation}
A graph $G$ is transformed into an auxiliary directed multigraph \( G' = (V', A') \) as follows. Initially, \( G' = (V', A') \) is a duplicate of \( G \). A dummy node \( v'_0 \) is added to the node set \( V' \) (\( V' = \{v'_0\} \cup V \)). This dummy node acts as an intermediate point for transitions between classes on each route. For example, the route $(v_0, a_{1}^1,\ldots, a_{k_1}^1,a_{1}^2,\ldots, a_{k_2}^2, \ldots, a_{1}^p,\ldots, a_{k_p}^p, v_0)$ in graph $G$ is transformed into the route $(v'_0, v_0, a_{1}^1,\ldots, a_{k_1}^1, v'_0, a_{1}^2,\ldots, a_{k_2}^2, v'_0, \ldots, v'_0, a_{1}^p,\ldots, a_{k_p}^p, v'_0) $ in graph $G'$. Here, $a_*^h$ represents a required arc in class $h\in P$, and $k_i$ is allowed to be $0$ for some $i \in P$.

Let $V_{t}^k$ be the set of tail vertices of arcs $A_r^{k}$ for each $k \in P$. Two sets of dummy arcs, \( A_f \) and \( A_t \), are then initialised as empty sets. For each vertex \( v \in V_t^1 \cup V_t^2 \cup ... \cup V_t^p \cup \{v_0\} \), arcs \( \{v, v'_0\} \) and \( \{v'_0, v\} \) are added to \( A_t \) and \( A_f \), respectively, with zero traversal time. These dummy arcs are also added to \( A' \).

Before presenting the \emph{Mixed Integer Linear Programming} (MILP) formulations, we need some additional notations. For a set $S \subseteq V'$, $A(S)$ denotes the set of arcs with both endpoints in $S$, while $\delta(S)$ represents the arcs with exactly one endpoint in $S$. Additionally, $\delta^{+}(S)$ and $\delta^{-}(S)$ represent the arcs leaving and entering $S$, respectively. Let \( A_r(S) = A(S) \cap A_r \), and similarly for \( \delta_r(S) \), \( \delta^{+}_r(S) \), and \( \delta^{-}_r(S) \). For each class \( k \in P \), let \( A^{k}(S) = A(S) \cap A^{k}_{r} \), and similarly for \( \delta^{k}(S) \), \( \delta^{+k}(S) \), and \( \delta^{-k}(S) \). For simplicity, \( \delta(v) \) is used instead of \( \delta(\{v\}) \).

\subsubsection{MILP model for the HDCARP-P}
\label{sub:milp-p}

The MILP model uses the following variables:
\begin{itemize}
\item $x_{a}^m \in \{0,1\}$ indicates whether vehicle $m \in M$ services arc $a \in A_r$.
\item $y_{ak}^m\in \mathbb{N}$ counts the number of times vehicle $m \in M$ deadheads through arc $a \in A'$ in class $k\in P$ in its chosen path. 
\item $t_k^m \in \mathbb{R}^+$ represents the service completion time of class $k \in P$ on route $m \in M$ (assigned to vehicle $m$). 
\item $r_k^m \in\{0,1\}$ indicates whether vehicle $m \in M$ services any required arcs in class $k \in P$.
\item $T_k\in \mathbb{R}^+$ represents the service completion time of class $k\in P$.
\end{itemize}

The MILP is then as follows:
\begingroup
\allowdisplaybreaks
\footnotesize
\begin{align}
\underset{k=1,...,p}{\textrm{lex-min}} \, \qquad T_k \label{eq:lex-obj}\\
\textrm{subject to} \qquad \;\; T_k &\geq t_k^m - N(1-r_k^m) \qquad m \in M; k\in P  \label{eq:mmc}\\
t_k^m &= t_{k-1}^m + \sum_{a \in A_r^k}s_ax_{a}^m + \sum_{a \in A}d_ay_{ak}^m \qquad m \in M; k\in P \label{eq:define-t}\\
t_0^m &= 0  \qquad m \in M  \label{eq:define-t-0}\\
\sum_{a \in A_r^k}x_{a}^m &\leq |A_r^k|r_k^m \qquad m \in M; k\in P \label{eq:service-route}\\
y_{\{v_0^\prime,v_0\} 1}^m &= 1 \qquad m \in M \label{eq:start-c0}\\
\sum_{a \in A_f}y_{ak}^m &= 1 \qquad m \in M; k\in P \label{eq:start-c}\\
y_{\{v^\prime_0,v_i\} k}^m &= y_{\{v_i,v^\prime_0\} k-1}^m \qquad m \in M; k \in P\setminus \{1\}; v_i \in \bigcup_{h=1}^{k-1} V_t^h \cup \{v_0\} \label{eq:class-link}\\
\sum_{m \in M} x_{a}^m &= 1 \qquad k\in P;a \in A_r^k \label{eq:required-arcs}\\
\sum_{k=1}^p\sum_{a \in A_r^k} q_ax_{a}^m &\leq Q \qquad m \in M \label{eq:lm-capacity}\\
\sum_{a \in \delta^+_k(v_i)}x_{a}^m  + \sum_{a \in \delta^+(v_i)}y_{ak}^m &= \sum_{a \in \delta^-_k(v_i)}x_{a}^m  + \sum_{a \in \delta^-(v_i)}y_{ak}^m \qquad m \in M; k\in P; v_i \in V^\prime \label{eq:lm-flow} \\
\sum_{a \in \delta^+_k(S)}x_{a}^m  + \sum_{a \in \delta^+(S)}y_{ak}^m &\geq x_{b}^m  \qquad m \in M; k\in P; S \subseteq V\setminus\{v_0\}; \forall b \in A_r^k(S) \label{eq:lm-subtour} \\
x_{a}^m \in \{0, 1\} &\qquad m \in M; k\in P; a \in A_r^k \label{eq:lm-x} \\
y_{ak}^m \in \mathbb{N}  &\qquad m \in M; k\in P;a \in A^\prime \label{eq:lm-y}\\
t_k^m \in \mathbb{R}^+ &\qquad m \in M ; k\in P \label{eq:lm-t} \\
r_k^m \in \{0, 1\} &\qquad m \in M ; k\in P \label{eq:lm-r} \\
T_k \in \mathbb{R}^+ &\qquad k\in P \label{eq:lm-z}
\end{align}
\endgroup

\normalsize
The objective function (\ref{eq:lex-obj}) represents the hierarchical objective, while constraints (\ref{eq:mmc}) ensure that the maximum completion time of class $k$ is greater than or equal to the completion time of that class on any route. Here, $N$ is a suitably large number. Constraints (\ref{eq:define-t}) and (\ref{eq:define-t-0}) are time-conservation constraints, ensuring route connectivity. Constraints (\ref{eq:service-route}) limit the number of arcs in class $k$ that can be serviced on each route. Constraints (\ref{eq:start-c0}), (\ref{eq:start-c}), and (\ref{eq:class-link}) define class transitions on each route using node $v'_0$. Constraints (\ref{eq:required-arcs}) and (\ref{eq:lm-capacity}) guarantee that each required arc is serviced exactly once and that vehicle capacity is not exceeded. Constraints (\ref{eq:lm-flow}) ensure that the number of times a vehicle departs from a node is equal to the number of times it arrives at that node for each class. Constraints (\ref{eq:lm-subtour}) state that if route $m$ services arc $b$, it must traverse any arc cutset that separates $b$ from the depot. The remaining constraints define the domains of the variables.

\subsubsection{MILP model for the HDCARP-U}
\label{sub:milp-u}

In the HDCARP-U, where class upgrading is permitted, the traversal order of classes within each route is unknown. The concept of hierarchy level is then introduced, representing the completion of a class’s service within a route. To be specific, in a route, hierarchy level 1 includes all required arcs from the departure up to the last required arc of class 1. Hierarchy level 2 includes the required arcs following hierarchy level 1 up to the last required arc of class 2, and so on. A hierarchy level may be an empty set. Transitions between hierarchy levels are represented by the dummy node \( v^{'}_{0} \).

The MILP model uses the following variables:
\begin{itemize}
\item $x_{ah}^m\in \{0,1\}$ indicates whether vehicle $m \in M$ services the required arc $a\in A_r$ in hierarchy level $h \in P$.
\item $y_{ah}^m\in \mathbb{N}$ counts the number of times vehicle $m \in M$ deadheads through arc $a \in A'$ in hierarchy level $h\in P$ in its chosen path.
\item $t_h^m \in \mathbb{R}^+$ represents the time at which vehicle $m \in M$ completes servicing all required arcs in hierarchy level $h \in P$. 
\item $r_{kh}^m\in \{0,1\}$ indicates whether hierarchy $h\in P$ contains any required arcs in class $k$ on route $m \in M$ (assigned to vehicle $m$).
\item $T_k\in \mathbb{R}^+$ represents the service completion time of class $k\in P$.
\end{itemize}

The MILP model for the HDCARP-U is described as follows:
\begingroup
\allowdisplaybreaks
\footnotesize
\begin{align}
\underset{k=1,...,p}{\textrm{lex-min}} \, \qquad T_k \label{eq:ulex-obj}\\
\textrm{subject to} \qquad \;\; T_k &\geq t_h^m - N(1-r_{kh}^m) \qquad m \in M; h,k\in P  \label{eq:ummc}\\
t_h^m &= t_{h-1}^m + \sum_{a \in A_r}s_ax_{ah}^m + \sum_{a \in A}d_ay_{ah}^m \qquad m \in M; h\in P \label{eq:udefine-t}\\
t_0^m &= 0  \qquad m \in M  \label{eq:udefine-t-0}\\
\sum_{a\in A_r^k}x_{ah}^m &\leq |A_r^k|r_{kh}^m\qquad m \in M; h, k \in P \label{eq:udefine-r}\\
y_{\{v_0^\prime,v_0\}1}^m &= 1 \qquad m \in M \label{eq:ustart-c0}\\
\sum_{a \in A_f}y_{ah}^m &= 1 \qquad m \in M; h\in P \label{eq:ustart-c}\\
y_{\{v^\prime_0,v_i\} h}^m &= y_{\{v_i,v^\prime_0\} h-1}^m \qquad m \in M; h \in P \setminus\{1\}; v_i \in V_t \cup \{v_0\} \label{eq:uclass-link}\\
\sum_{m \in M}\sum_{h=1}^p x_{ah}^m &= 1 \qquad a \in A_r \label{eq:urequired-arcs}\\
\sum_{h=1}^p\sum_{a \in A_r}q_a x_{ah}^m &\leq Q \qquad m \in M \label{eq:ulm-capacity}\\
\sum_{a \in \delta^+_r(v_i)}x_{ah}^m  + \sum_{a \in \delta^+(v_i)}y_{ah}^m &= \sum_{a \in \delta^-_r(v_i)}x_{ah}^m  + \sum_{a \in \delta^-(v_i)}y_{ah}^m \qquad m \in M; h\in P; v_i \in V^\prime \label{eq:ulm-flow} \\
\sum_{a \in \delta^+_r(S)}x_{ah}^m  + \sum_{a \in \delta^+(S)}y_{ah}^m &\geq x_{bh}^m  \qquad  m \in M; h\in P; S \subseteq V\setminus \{v_0\}; \forall b \in A_r(S) \label{eq:ulm-subtour} \\
x_{ah}^m \in \{0, 1\} &\qquad m \in M; h\in P; a \in A_r \label{eq:ulm-x} \\
y_{ah}^m \in \mathbb{N}  &\qquad m \in M; h\in P; a \in A^\prime \label{eq:ulm-y}\\
t_h^m \in \mathbb{R}^+ &\qquad m \in M; h\in P  \label{eq:ulm-t} \\
r_{kh}^m \in \{0, 1\} &\qquad m \in M; h, k \in P  \label{eq:ulm-r} \\
T_k \in \mathbb{R}^+ &\qquad k\in P \label{eq:ulm-z}
\end{align}
\endgroup

\normalsize
Constraints (\ref{eq:ummc}) ensure that the maximum completion time of a class is greater than or equal to the completion time of any hierarchy level on any route that includes at least one required arc of that class. Constraints (\ref{eq:udefine-t}), (\ref{eq:urequired-arcs}), and (\ref{eq:ulm-subtour}) are analogous to constraints (\ref{eq:define-t}), (\ref{eq:required-arcs}), and (\ref{eq:lm-subtour}) of the HDCARP-P model, respectively, with the exception that required arcs can be serviced at any hierarchy level. Constraints (\ref{eq:udefine-r}) limit the maximum number of required arcs of class $k$ that can be serviced in a single hierarchy level on each route. The remaining constraints and objective function are straightforward.

\subsubsection{Branch-and-cut algorithm}

The models are first solved without the connectivity constraints ((\ref{eq:lm-subtour}) for the HDCARP-P and (\ref{eq:ulm-subtour} for the HDCARP-U). Violated connectivity inequalities are then identified and added to the current MILP, which is reoptimised. This process is repeated until all connectivity constraints are satisfied. If fractional variables remain, branching generates two new sub-problems; if all variables are integers, another sub-problem is explored. Violated connectivity inequalities are identified using a heuristic. This heuristic computes the connected components in a graph constructed from the current solution and checks whether these connected components satisfy the connectivity constraints. For further details, see \cite{https://doi.org/10.1002/net.21525}.

\subsection{Proximal Policy Optimization (PPO)}
\label{appendix:ppo}
The heart of PPO is a surrogate objective function that optimizes the new policy \( \pi_\theta \) while maintaining proximity to an older reference policy \( \pi_{\theta_\text{old}} \). The key to PPO's stability is its clipping mechanism, which ensures that the probability ratio between the new and old policies does not deviate excessively. This is formalized in the following objective function:
\begin{align*}
\mathcal{L}_{\text{CLIP}}(\theta) = \mathbb{E}_{x \sim P(x)} \Bigg[ \ 
    \mathbb{E}_{\mathbf{\alpha} \sim \pi_{\theta_\text{old}}(\mathbf{\alpha} \mid x)} \Bigg[ \ 
        \min \Bigg( \ 
            & \frac{\pi_\theta(\mathbf{\alpha} \mid x)}{\pi_{\theta_\text{old}}(\mathbf{\alpha} \mid x)} \, A^{\pi_{\theta_\text{old}}}(x, \mathbf{\alpha}), \\
            & \text{clip} \left( 
                \frac{\pi_\theta(\mathbf{\alpha} \mid x)}{\pi_{\theta_\text{old}}(\mathbf{\alpha} \mid x)}, 
                1 - \epsilon, \ 
                1 + \epsilon 
              \right) \, A^{\pi_{\theta_\text{old}}}(x, \mathbf{\alpha}) 
        \Bigg) \ 
    \Bigg] \ 
\Bigg],
\end{align*}
where \( A^{\pi_{\theta_\text{old}}}(x, \mathbf{\alpha}) \) is the advantage function at state \( x \) and action \( \mathbf{\alpha} \) under the reference policy \( \pi_{\theta_\text{old}} \). It is defined as the difference between the actual reward \( R(x, \mathbf{\alpha}) \) and the expected value of the state \( V_{\phi}(x) \), specifically:

\[
A^{\pi_{\theta_\text{old}}}(x, \mathbf{\alpha}) = R(x, \mathbf{\alpha}) - V_{\phi}(x).
\]

The advantage function \( A^{\pi_{\theta_\text{old}}}(x, \mathbf{\alpha}) \) plays a crucial role in PPO as it determines the direction and magnitude of the policy update. The value function \( V_{\phi}(x) \) is trained to minimize the mean squared error between the estimated value and the actual return:

\[
\mathcal{L}_V(\phi) = \mathbb{E}_{x \sim P(x)} \left[ \left( R(x, \mathbf{\alpha}) - V_{\phi}(x) \right)^2 \right].
\]

During each optimization step, PPO updates both the policy parameters \( \theta \) and the value function parameters \( \phi \). The overall loss function in PPO combines the clipped surrogate objective \( \mathcal{L}_{\text{CLIP}} \) and the value function loss \( \mathcal{L}_V(\phi) \), often with an additional weighting factor \( \beta \) to balance their contributions:

\[
\mathcal{L}_{\text{PPO}} = \mathcal{L}_{\text{CLIP}} + \beta \mathcal{L}_V(\phi).
\]

By constraining the updates to the policy in this way, PPO ensures more stable and consistent improvements in performance, making it a highly effective algorithm for complex reinforcement learning tasks.

\subsection{Policy network}
\label{appendix:policynetwork}

\subsubsection{Encoder}

\paragraph{Initialization of embeddings}

\begin{itemize}
    \item \textbf{Input features preparation:}

    The input features for each required arc in the graph include several attributes:

    \[
    \begin{aligned}
    q_i &: \text{Demand of arc } i \\
    p_i &: \text{Priority classes of arc } i \\
    s_i &: \text{Servicing time of arc } i \\
    d_i &: \text{Traversal time of arc } i
    \end{aligned}
    \]

    These attributes are combined into a feature matrix $F \in \mathbb{R}^{N \times 4}$ for all required arcs except the depot:

    \[
    F_i = [q_i, p_i, s_i, d_i], \quad i = 1, \ldots, N
    \]

    \item \textbf{Arc embedding calculation:}

    The feature matrix $F$ is transformed into embeddings using a linear transformation:

    \[
    E = F W + b,
    \]

    where:
    \begin{itemize}
        \item $W \in \mathbb{R}^{4 \times \text{embed\_dim}}$ is the weight matrix.
        \item $b \in \mathbb{R}^{\text{embed\_dim}}$ is the bias vector.
        \item $E \in \mathbb{R}^{N \times \text{embed\_dim}}$ represents the arc embeddings.
    \end{itemize}

    \item \textbf{Depot embedding calculation:}

    The depot is represented using a zero vector, and its embedding is computed separately:

    \[
    F_{\text{depot}} = [0, 0, 0, 0].
    \]

    The embedding for the depot is given by:

    \[
    E_{\text{depot}} = F_{\text{depot}} W_{\text{depot}} + b_{\text{depot}},
    \]

    where:
    \begin{itemize}
        \item $W_{\text{depot}} \in \mathbb{R}^{4 \times \text{embed\_dim}}$ is the weight matrix for the depot.
        \item $b_{\text{depot}} \in \mathbb{R}^{\text{embed\_dim}}$ is the bias vector.
        \item $E_{\text{depot}} \in \mathbb{R}^{1 \times \text{embed\_dim}}$ is the depot embedding
    \end{itemize}

    \item \textbf{Concatenation of depot and arc embeddings:}

    The depot embedding is concatenated with the arc embeddings to form the complete set of embeddings:

    \[
    E_{\text{all}} = \begin{bmatrix} E_{\text{depot}} \\ E \end{bmatrix},
    \]

    where $E_{\text{all}} \in \mathbb{R}^{(N+1) \times \text{embed\_dim}}$.
\end{itemize}

\paragraph{Graph attention network for HDCARP}
\label{appendix:GNN}
The Graph Attention Network (GAT) learns arc representations by applying multiple layers, where each layer $l$ updates the input embeddings $x^{(l)}$ using an adjacency matrix that encodes the distances between arcs:
\[
x^{(l+1)} = A_l(x^{(l)}, \mathcal{D}),
\]
where:
\begin{itemize}
    \item $x^{(l)} \in \mathbb{R}^{B \times S \times D}$ is the input to layer $l$, with $B$ as the batch size, $S$ as the number of arcs, and $\mathcal{D}$ as the embedding dimension.
    \item $\mathcal{D} \in \mathbb{R}^{B \times S \times S}$ is the adjacency matrix.
    \item $A_l$ comprises a skip connection, multi-head attention, feed-forward network (MLP), and batch normalization.
\end{itemize}

\paragraph{Layer operations}

\begin{itemize}
    \item \textbf{Multi-Head attention with skip connection:}

The input embeddings $x^{(l)}$ are processed using a multi-head attention mechanism and a skip connection to preserve the original input:

\[
h^{(l)} = x^{(l)} + \text{MHA}(x^{(l)}, \mathcal{D}),
\]

where the multi-head attention (MHA) mechanism involves several steps:

\begin{itemize}
    \item \textbf{Linear transformations:} The input $x^{(l)}$ is linearly transformed to obtain queries ($Q$), keys ($K$), and values ($V$) for each head:
    \[
    Q = x^{(l)}W_q, \quad K = x^{(l)}W_k, \quad V = x^{(l)}W_v,
    \]
    where $W_q, W_k, W_v \in \mathbb{R}^{D \times d_k}$ are learnable weight matrices, and $d_k$ is the dimension of each head.

    \item \textbf{Scaled dot-product attention and masking with adjacency matrix for each head:} For each attention head $i$, the scaled dot-product attention is calculated as:
    \[
    \text{head}_i=
    \text{Attention}(Q_i, K_i, V_i) = \text{Softmax}\left(\frac{Q_i K_i^T}{\sqrt{d_k}} \odot \mathcal{D}\right) V_i,
    \]
    where $Q_i, K_i, V_i$ are the query, key, and value matrices for head $i$, derived by slicing $Q, K, V$ respectively.

    \item \textbf{Concatenation of heads:} The outputs from all heads are concatenated to form the multi-head attention output:
    \[
    \text{MHA}(Q, K, V) = \text{Concat}(\text{head}_1, \text{head}_2, \ldots, \text{head}_H) W^O,
    \]
    where $\text{head}_i$ is the output of the $i$-th attention head; and $W^O \in \mathbb{R}^{H \cdot d_k \times D}$ is a learnable weight matrix that combines the concatenated outputs of all heads back into the original embedding dimension $\mathcal{D}$.
\end{itemize}

    \item \textbf{Feed-forward network (MLP):}

    The attention output is fed into a feed-forward network:

    \[
    \text{MLP}(h^{(l)}) = W_2 \cdot \text{ReLU}(W_1 \cdot h^{(l)} + b_1) + b_2.
    \]

    \item \textbf{Batch normalization:}
    
    The final output of the layer is normalized:

    \[
    x^{(l+1)} = \text{BatchNorm}(\text{FF}(h^{(l)})).
    \]

    Batch normalization is defined as:

    \[
    \text{BatchNorm}(x) = \gamma \frac{x - \mu}{\sqrt{\sigma^2 + \epsilon}} + \beta,
    \]

    where $\mu$ and $\sigma^2$ are the mean and variance of the batch, $\gamma$ and $\beta$ are learnable parameters, and $\epsilon$ is a small constant for numerical stability.
\end{itemize}

\end{document}